\documentclass[conference]{IEEEtran}
\usepackage{etoolbox}
\makeatletter
\patchcmd{\@makecaption}
  {\scshape}
  {}
  {}
  {}
\makeatletter
\patchcmd{\@makecaption}
  {\\}
  {.\ }
  {}
  {}
\makeatother
\def\tablename{Table}

\IEEEoverridecommandlockouts
\usepackage{cite}
\usepackage{amsmath,amssymb,amsfonts}
\usepackage{graphicx}
\usepackage{textcomp}
\usepackage{xcolor}

\usepackage{helvet}
\usepackage{courier}
\usepackage{subfigure}
\usepackage{amssymb}
\usepackage{enumitem}
\usepackage{amsmath,bm}
\usepackage{color}

\usepackage{algorithmicx,algorithm}
\usepackage[noend]{algpseudocode} 

\usepackage{changes}
\usepackage{setspace}
\usepackage{graphbox}
\usepackage{hyperref}
\usepackage{makecell}
\usepackage{diagbox}
\usepackage{multirow}
\usepackage{extarrows}

\newtheorem{define}{Definition}

\newtheorem{remark}{Remark}

\def\BibTeX{{\rm B\kern-.05em{\sc i\kern-.025em b}\kern-.08em
    T\kern-.1667em\lower.7ex\hbox{E}\kern-.125emX}}
\begin{document}

\def\Algname{Interactive System-wise Anomaly Detection}
\def\Algnameabbr{InterSAD}

\title{Interactive System-wise Anomaly Detection}
% through System Embedding Neutralization and Experience Replay Mechanism

% \title{Towards Anomaly Detection in Markov Decision Process}

% Guanchu Wang, Ninghao Liu, Daochen Zha, and Xia Hu

\author{\IEEEauthorblockN{Guanchu Wang} 
% \IEEEauthorblockA{\textit{dept. name of organization (of Aff.)} \\
\textit{Rice University} \\
% Rice University \\
\texttt{guanchu.wang@rice.edu}
\and
\IEEEauthorblockN{Ninghao Liu} 
% \IEEEauthorblockA{\textit{dept. name of organization (of Aff.)} \\
\textit{University of Georgia} \\
\texttt{ninghao.liu@uga.edu} 
\and
\IEEEauthorblockN{Daochen Zha} 
% \IEEEauthorblockA{\textit{dept. name of organization (of Aff.)} \\
\textit{Rice University} \\
\texttt{daochen.zha@rice.edu}
\and
\IEEEauthorblockN{Xia Hu} 
% \IEEEauthorblockA{\textit{dept. name of organization (of Aff.)} \\
\textit{Rice University} \\
\texttt{xia.hu@rice.edu}
}

\maketitle

\noindent
\begin{abstract}
Anomaly detection, where data instances are discovered containing feature patterns different from the majority, plays a fundamental role in various applications. However, it is challenging for existing methods to handle the scenarios where the instances are systems whose characteristics are not readily observed as data. Appropriate interactions are needed to interact with the systems and identify those with abnormal responses. Detecting system-wise anomalies is a challenging task due to several reasons including: how to formally define the system-wise anomaly detection problem; how to find the effective activation signal for interacting with systems to progressively collect the data and learn the detector; how to guarantee stable training in such a non-stationary scenario with real-time interactions?
To address the challenges, we propose \Algnameabbr{}~(\Algname{}). Specifically, first, we adopt Markov decision process to model the interactive systems, and define anomalous systems as anomalous transition and anomalous reward systems. Then, we develop an end-to-end approach which includes an encoder-decoder module that learns system embeddings, and a policy network to generate effective activation for separating embeddings of normal and anomaly systems. Finally, we design a training method to stabilize the learning process, which includes a replay buffer to store historical interaction data and allow them to be re-sampled.
Experiments on two benchmark environments, including identifying the anomalous robotic systems and detecting user data poisoning in recommendation models, demonstrate the superiority of \Algnameabbr{} compared with state-of-the-art baselines methods.
The source code is available at \href{https://anonymous.4open.science/r/InterSAD-E194/}{\texttt{https://anonymous.4open.science/r/InterSAD-E194/}}.

%Existing work is incapable to formulate the anomaly detection problem of interactive scenarios where the behaviors of each system could not be explicitly observed as data unless an activation signal is provided.
% (i) existing work lacks formal definitions for interactive systems-wise anomaly detection;
% (ii) it relies on real-time interaction with the systems to progressively collect the data for learning the detector, which contradicts the assumptions of traditional anomaly detection that operate on static datasets; 
% (iii) the data collected from real-time interaction with the systems has non-stationary and noisy distribution, which can terribly affect the stability of training procedure.

\end{abstract}

% \begin{IEEEkeywords}
% component, formatting, style, styling, insert
% \end{IEEEkeywords}

\section{Introduction}

% Anomaly Detection

Anomaly detection is a crucial task in various application domains, e.g. fraud detection in finance~\cite{ahmed2016survey}, intrusion detection in network security~\cite{garcia2009anomaly} and damage detection for electronic systems~\cite{fu2006finding, keogh2002finding, yankov2008disk}. % or sensor network, industrial
Anomalies, or anomalous instances, usually refer to the minority of data points or patterns that do not conform to the characteristics of the majority~\cite{zhao2019pyod, lai2020tods, li2020pyodds}.
It is necessary to identify anomalous instances because they are often associated with malicious activities or breakdown of a system, and can provide more actionable information than normal instances towards system improvement.
However, it is difficult for traditional detection methods to formulate the scenario where instances are neither data points nor time-series patterns but \emph{systems}, since the behaviors of each system could \emph{not} be explicitly observed as data unless an activation signal is provided. Due to the faulted or malicious inner characteristics, anomaly systems tend to behave or response differently from others.
In this work, we investigate this problem and formulate it as \emph{system-wise anomaly detection}, where the systems refer to a group of objects or entities that receive activation as input and produce reactions as output.

%e.g. designing input signals to activate the machines so that the broken ones can be identified according to their output signals.
% of the same type 

\begin{figure}
% \setlength{\abovecaptionskip}{3mm}
% \setlength{\belowcaptionskip}{-0.3cm}
% \subfigbottomskip=-5mm
% \subfigcapskip=-2mm
\centering
\begin{minipage}[b]{0.21\textwidth}
\centering
\subfigure[]{
\centering
\includegraphics[width=0.8\textwidth]{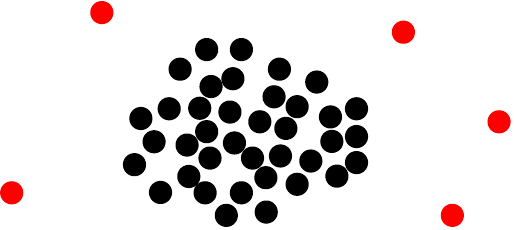}}\\
\subfigure[]{
\centering
\includegraphics[width=1.0\textwidth]{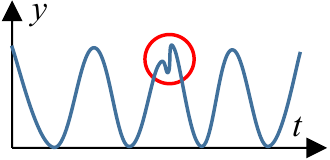}
}
\end{minipage}
\begin{minipage}[b]{0.25\textwidth}
\centering
\subfigure[]{
\centering
\includegraphics[width=1.0\textwidth]{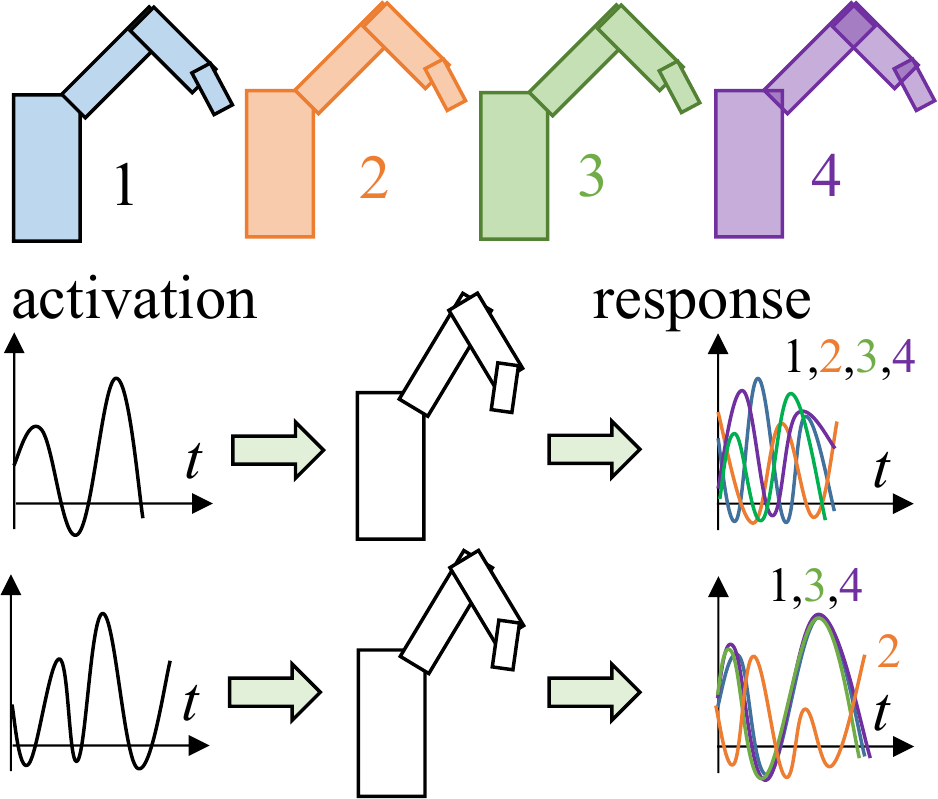}
}
\end{minipage}
\caption{
(a) Point-wise anomaly. 
(b) Pattern-wise anomaly in time-series.
(c) System-wise anomaly. Upper: 4 robotic systems (system 2 is anomalous). 
Middle: inconsistent outputs in response to a random activation.
Lower: consistent outputs from normal systems in response to a well-designed activation which can isolate the anomalous system. 
}
\label{fig:anomalous_example_robot}
\vspace{-10pt}
\end{figure}

\begin{figure}
% \setlength{\abovecaptionskip}{-2mm}
% \setlength{\belowcaptionskip}{-0.5cm}
% \subfigcapskip=-2mm
\centering
\begin{minipage}{0.62\linewidth}
\centering
\subfigure[Interactions with normal users.]{
\centering
\includegraphics[width=1.0\textwidth]{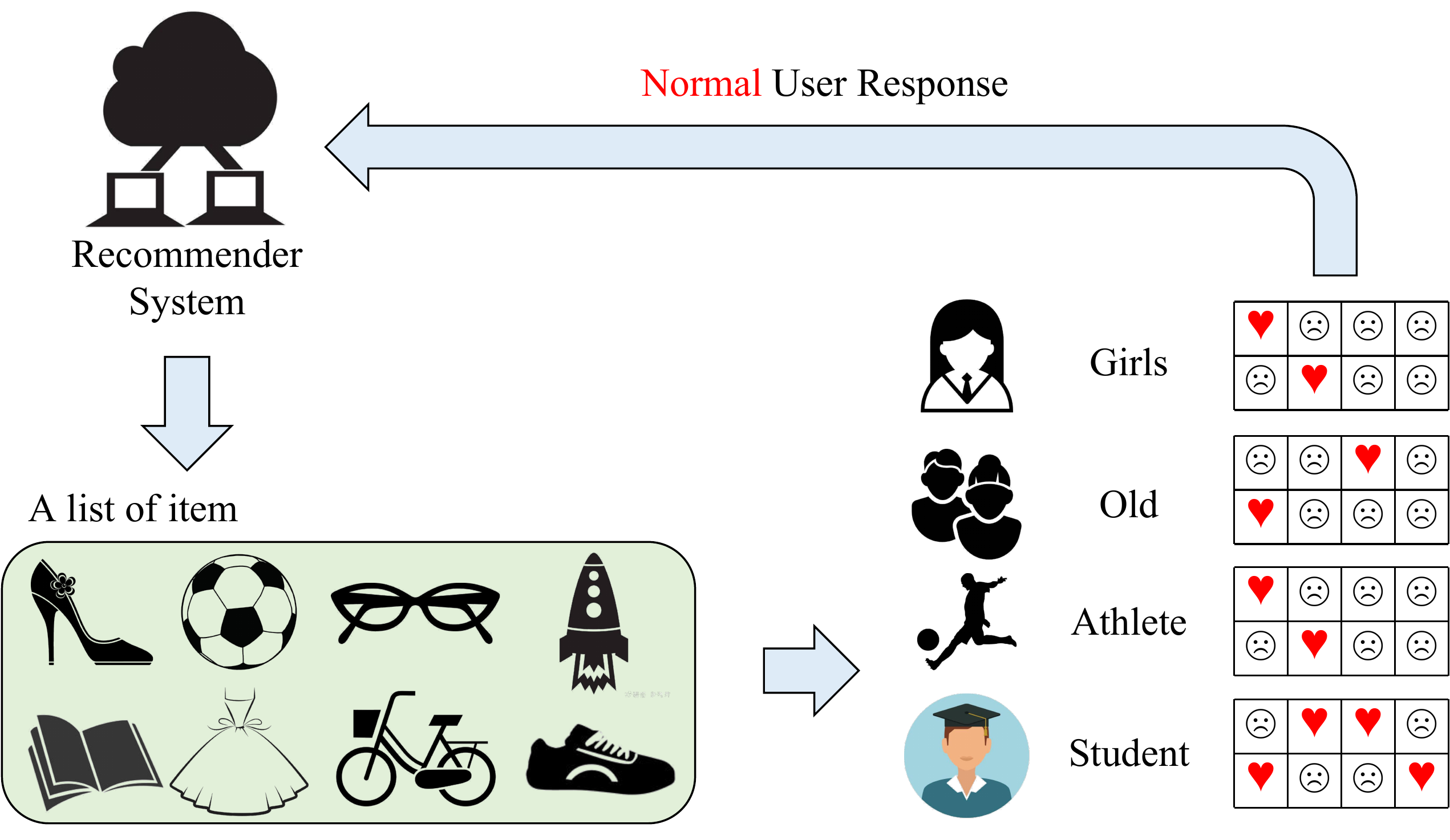}
}
\end{minipage}
\text{  }
\begin{minipage}{0.28\linewidth}
\centering
\subfigure[Malicious users.]{
\centering
\raisebox{0.05\height}{\includegraphics[width=1.0\textwidth]{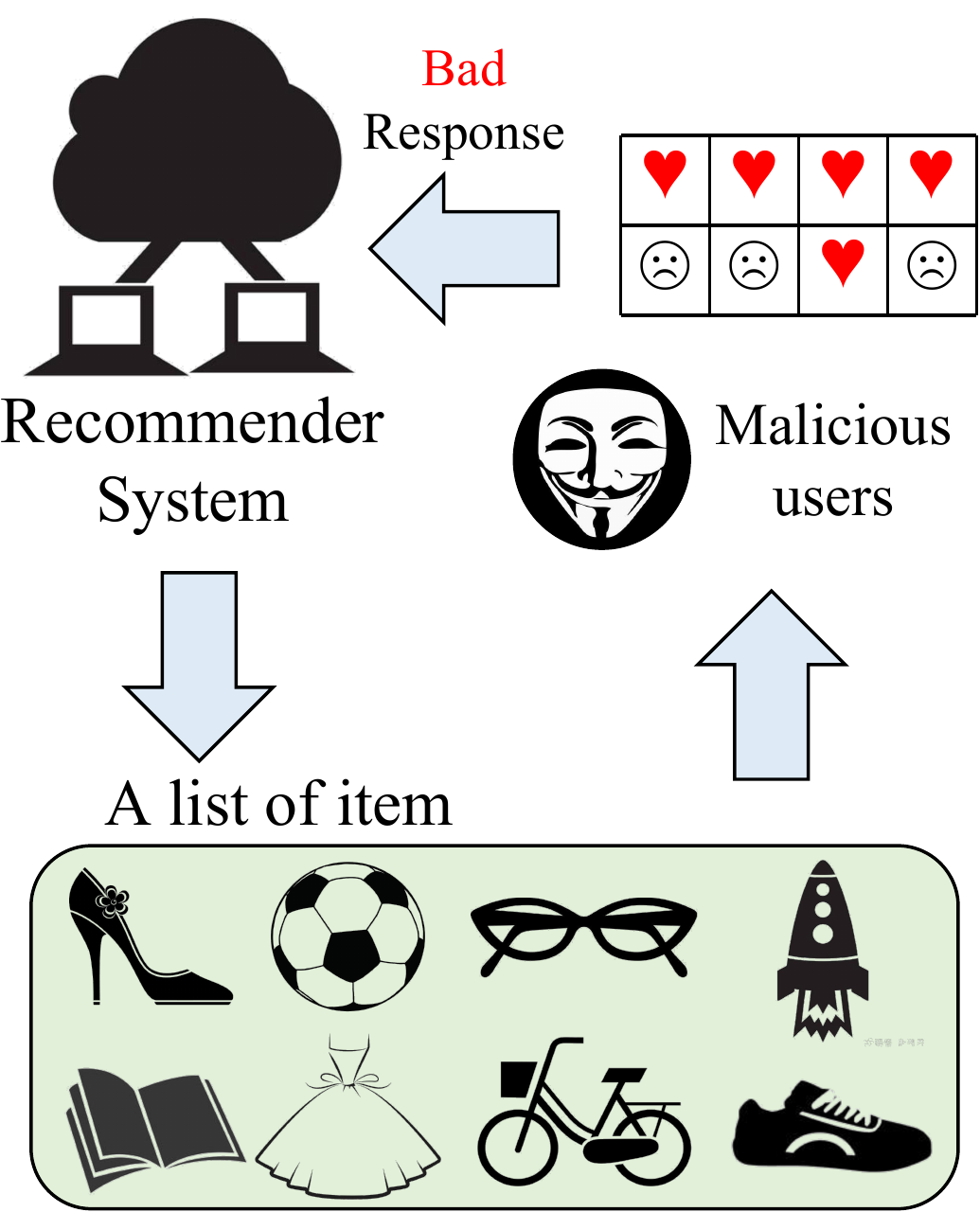}}
}
\end{minipage}
\caption{(a) The recommender system collects users' responses in the interaction with normal users to learn their preference.
(b) Anomalous users maliciously attack the system by providing fake user preference.}
\label{fig:anomalous_example_recsys}
\vspace{-8pt}
\end{figure}

Different from traditional anomaly detection scenarios~\cite{chalapathy2019deep}, such as point-wise or time-series anomaly detection shown in Figures~\ref{fig:anomalous_example_robot}~(a) and (b), system-wise anomaly detection aims to identify systems which have significantly different behaviors from the majority.
Here we provide two examples of system-wise anomaly detection to show its application in robotic and recommender systems, respectively.
First, as shown in Figure~\ref{fig:anomalous_example_robot}~(c), robots can be regarded as systems that respond to an external activation at different time points~\cite{brockman2016openai}.
Some robotic systems may behave falsely or differently from the majority in response to the activation, implying flawed components in hardware. 
Second, as shown in Figure~\ref{fig:anomalous_example_recsys}~(a), a recommendation model learns the preference of users based on their historically clicked items~\cite{shi2019virtual}.
However, there exist malicious users who strategically click the items that contradict those of the user majority to prevent the recommendation model from learning the preference of majority users~\cite{wilson2013power, aktukmak2019quick}, as shown in Figure~\ref{fig:anomalous_example_recsys}~(b).
The detection of such malicious users can be formulated as system-wise anomaly detection, where each user can be seen as a system that accepts items and outputs clicking rate.
The detection of such anomalous robotic systems could prevent bigger losses in various applications in the future. 
%Beyond the above application scenarios, system-wise anomaly detection could also benefit many other tasks since the objects in different domains can always be modeled as black-box systems with specific input and output.
However, there lacks an effective detection algorithm or even a formal definition concerning anomalous systems to the best of our knowledge.
To bridge this gap, we aim to provide a unified formulation of anomalous systems and propose an effective framework for system-wise anomaly detection.

% different from conventional anomaly detection scenarios, such as point-wise, contextual, or collective anomaly detection, which operate on static datasets with fixed instance representations~\cite{chalapathy2019deep},
% MDP describes an interactive process with dynamic state representations.
% As a result, traditional algorithms will usually fail in the context of MDP. Even though some prior work has studied anomaly detection for multi-armed bandit~\cite{zhuang2017identifying,ban2020generic}, they assume that there is no state transition in the environment, which restricts their applicability in many scenarios, such as anomalous hard-ware detection.

It is non-trivial to achieve the goal due to the following challenges.
First, system-wise anomaly detection relies on real-time interactions with the systems, to be more concrete, providing activation to and collecting responses from the systems. It is significantly different from traditional scenarios where data instances or patterns are readily given.
The main challenge is how to learn a generator to provide effective activation to the systems in order to successfully trigger consistent response from normal systems to isolate anomalous systems, as illustrated at the bottom of Figure~\ref{fig:anomalous_example_robot} (c).
The generator should be well-trained because random activations may cause inconsistent responses from normal systems such that anomalous systems are less distinguishable, as shown in the middle of Figure~\ref{fig:anomalous_example_robot} (c).
Furthermore, different from static dataset-based training, where the training process is stable due to the stationary distribution of training data, system-wise anomaly detection depends on real-time interactions with each system, where the real-time data collected from the systems have non-stationary and noisy distribution.
The non-stationary distributed training data can terribly affect the stability of training process or even lead to the failure of convergence.

% we need to provide activation to the systems and collect their responses (or behaviors) as data, so that the anomalous system can be identified 

In this work, we propose \Algnameabbr{} (\Algname{}) to detect anomalous systems. Specifically, each system is modeled as a Markov decision process (MDP) that interacts with an activation signal. The abnormal characteristics of a system could manifest through its state and reward function, so we categorize the detection of anomalous systems into anomalous transition and anomalous reward systems.
Meanwhile, to tackle the first challenge, we propose to learn a neural network to generate the activation, which is termed as the \emph{policy} in the remaining parts.
To encourage normal systems to have consistent response, we optimize the policy to neutralize the embeddings of different systems, where system embeddings are provided by an encoder-decoder model that is learned simultaneously with the policy. 
To address the second challenge, we employ experience replay mechanism (ERM) to stabilize the training process. Specifically, the real-time responses collected from systems are preserved in a replay buffer. The encoder-decoder is updated with the data sampled from the buffer, which encourages more stable distribution than the data collected from real-time interactions.
The contributions are summarized as follows:

\begin{itemize}[leftmargin=11pt, topsep=0pt]

% \item We formalize the problem of system-wise anomaly detection based on the concept of Markov decision process.

\item We propose \Algnameabbr{} which learns a policy and encoder-decoder for end-to-end system-wise anomaly detection. 
%Meanwhile, we consider two types of anomalous systems, i.e., anomalous transition and anomalous reward systems.
% anomalous reward and transitions.

\item We employ ERM to stabilize the training process so that \Algnameabbr{} can be trained based on the non-stationary data collected from the interaction with the systems.

\item The effectiveness of \Algnameabbr{} is demonstrated with experiment results on anomalous robotic systems identification and user attack detection in recommendation models.

% \item We introduce two practical instances of \Algnameabbr{} by learning the embedding network according to the reward sequences and trajectories, which can abstract the trajectories into lower-dimensional embedding space.

\end{itemize}

\section{Preliminaries}
In this section, we introduce the notations, background and problem definition. 
Specifically, we briefly introduce Markov decision process that is adopted to model interactive systems.
% formalize the interaction with systems. 
We also define the problem of system-wise anomaly detection from two perspectives including anomalous reward and anomalous transition.

% In this section, we provide a background of the Markov Decision Process~(MDP), 
% Then we discuss the limitations of anomaly detection algorithms for multi-armed bandits. Finally, we formally define the problem of anomaly detection in MDP.

\subsection{Markov Decision Process}

Markov decision process (MDP) describes a discrete-time stochastic process represented by state space $\mathcal{S}$, activation space $\mathcal{A}$, reward space $\mathbb{R}$ and state transition $\mathcal{S} \times \mathcal{A} \to \mathcal{S}$.
Here, we employ the MDP model to characterize the interactive systems. 
Specifically, in each time step $t$, a system is in state $s_t \in \mathcal{S}$, receives an activation $a_t \in \mathcal{A}$, transits from $s_t$ to $s_{t+1}$, and generates a reward signal $r_t \in \mathbb{R}$.
The state transition of MDP can be widely applied in various domains to model system behaviors in response to external activation.
For example, in recommender systems~\cite{zheng2018drn}, a user receives a list of recommended items $a_t$, gives response $r_t$ to the recommendation, and could have a shift of interest $s_t \to s_{t+1}$.
In robotic control~\cite{kober2013reinforcement}, a robot makes movements $s_t \to s_{t+1}$ in response to an external electrical activation $a_t$.
During interactions, the state transitions of each system along the discrete timeline $t = 0, 1, \cdots, T$ result in a trajectory $\boldsymbol{\tau} = \{ s_0, a_0, \cdots, s_{T-1}, a_{T-1} \}$ and a reward sequence $\boldsymbol{r} = [r_0, \cdots, r_{T-1}]$, where $T$ denotes the maximum time step of each interaction. 
These information reflect the characteristics of a system, thus is used for measuring its abnormality degree.

\begin{figure*}[t]
\setlength{\abovecaptionskip}{0.15cm}
\setlength{\belowcaptionskip}{-0.5cm}
\centering
\includegraphics[width=0.9\textwidth]{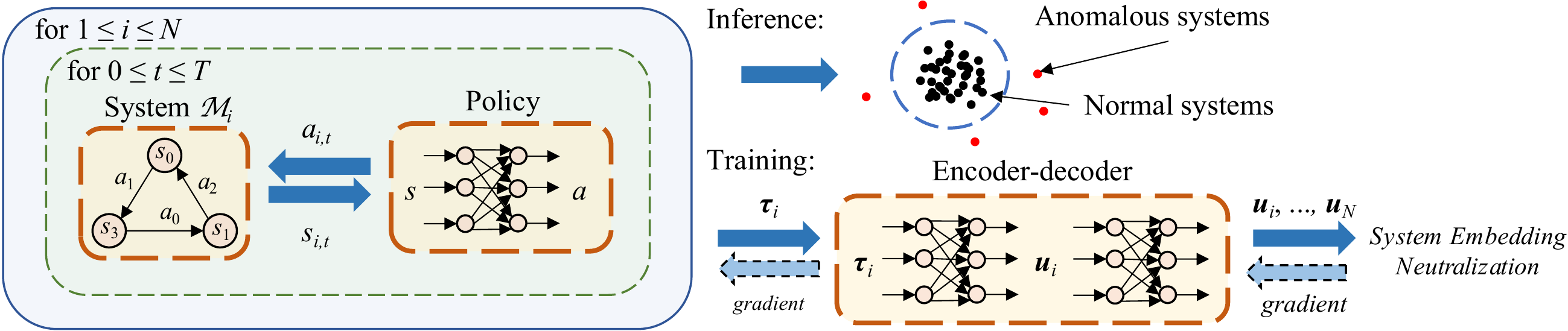}
\caption{An overview of \Algnameabbr{} framework. A policy strategically selects actions to interact with the systems; an encoder-decoder model learns the embedding of the systems with the collected trajectories. }
\label{fig:SEN_block}
\end{figure*}

\subsection{Problem Formulation}

To formally define the problem of system-wise anomaly detection, we consider $N$ MDP systems $\mathcal{M}_1, \mathcal{M}_2, \cdots, \mathcal{M}_N$ with identical state space $\mathcal{S}$ and activation space $\mathcal{A}$.
Without loss of generality, each system $\mathcal{M}_i$ accepts activation $\boldsymbol{a}_i = [a_{i,0}, \cdots, a_{i,T-1}]$ to outputs a tuple of trajectory $\boldsymbol{\tau}_i = [s_{i,0}, a_{i,0}, \cdots, s_{T-1}, a_{T-1}]$ and reward sequence $\boldsymbol{r}_i = [r_{i,0}, \cdots, r_{i,T-1}]$, which can be characterized by
\begin{equation}
\begin{aligned}
\boldsymbol{a}_i \xLongrightarrow{\text{input}} \mathcal{M}_i \xLongrightarrow{\text{output}} (\boldsymbol{\tau}_i, \boldsymbol{r}_i).
\end{aligned}
\end{equation}

In anomaly detection, we want to identify those systems with rare and distinctive characteristics compared with the majority. 
However, different from the point-wise and time-series anomaly detection where data is already given, the behaviors of systems are not readily available unless activations are given. In this work, we provide identical activation to all of the systems, where anomalous systems refers to those having significantly different reactions compared with other systems.
The formal definition is given as below.
%We propose the concept of anomalous system based on that of \emph{anomaly}, which refers to the instance that is distinctive compared with the majority.
%and the interpretation of Definition~\ref{def:anomalous_sys} in Remarks~\ref{rk:3}-\ref{rk:3}.

\begin{define}[Anomalous System]
\label{def:anomalous_sys}
Given the activation $\boldsymbol{a}_i$ to interact with the systems, i.e. $\boldsymbol{a}_i \!\xLongrightarrow{\text{input}}\! \mathcal{M}_i \!\xLongrightarrow{\text{output}}\! (\boldsymbol{\tau}_i, \boldsymbol{r}_i)$ for $1 \leq i \leq N$, a system $\mathcal{M}_i$ is defined as an \emph{anomalous transition system} if it has considerable different state transitions in contrast with other systems;
similarly, a system $\mathcal{M}_i$ is defined as an \emph{anomalous reward system} if it responds with significantly different reward sequence patterns compared with those of most other systems. 
\end{define}
%, i.e. the features of $\boldsymbol{r}_i$ is quite different from those of $\boldsymbol{r}_j$ for $1 \leq j \neq i \leq N$
%, i.e. the feature of $\boldsymbol{\tau}_i$ significantly differs from that of $\boldsymbol{\tau}_j$ for $1 \leq j \neq i \leq N$

\begin{remark}
\label{rk:1}
An anomalous transition system refers to the system where anomaly patterns exist in its state transitions $s_{i,t}, a_t \to s_{i,t+1}$ during the interactions.
For example, given the same external activation, an anomalous robot will have different response compared with normal robots.
An anomalous reward system refers to the system where anomaly patterns exist in its generated reward sequence $\boldsymbol{r}_i$. 
For example, anomalous users in recommender system tend to click the items that are different from those of normal users, thus returning different reward signals. 
\end{remark}

\begin{remark}
\label{rk:2}
Unlike traditional anomaly detection tasks where the features of each instance are already given in the dataset to learn a detector, the detection of anomalous MDP systems requires an activation signal $\boldsymbol{a}$ to trigger each system and collect its responses as features. Such an activation is not readily available, and we develop an algorithm to find the appropriate activation $\boldsymbol{a}$ for system $\{ \mathcal{M}_i \}_{1 \leq i \leq N}$ to collect reward and trajectory sequences, respectively.
\end{remark}

\begin{remark}
\label{rk:3}
For each system $\mathcal{M}_i$, its anomaly score in transitions and rewards is measured as $\phi(\boldsymbol{\tau}_i; \{ \boldsymbol{\tau}_i \}_{1 \leq i \leq N})$ and $\phi(\boldsymbol{r}_i; \{ \boldsymbol{r}_i \}_{1 \leq i \leq N})$, respectively. Here $\phi(\cdot; \cdot)$ is the scoring function, where the first operand denotes the instance of interest, and the second operand includes all instances as the context for contrast.
In this work, we do not constrain the scoring function to be any specific type, so $\phi(\cdot; \cdot)$ can be constructed from commonly used anomaly detection models or algorithms, such as the angle-based outlier detector~\cite{kriegel2008angle}, isolation forest~\cite{Liu2012} or density-based outlier detector~\cite{Breunig2000} and one-class classification model~\cite{Manevitz2002}.
\end{remark}
%Here $\phi(\boldsymbol{x}_i; \{\boldsymbol{x}_i\}_{1 \leq i \leq N})$ denotes a $\mathcal{X}^N \in \mathbb{R}$ mapping to indicate the novelty of $x_i$ for $x_i \in \{\boldsymbol{x}_i\}_{1 \leq i \leq N}$.

From the discussion above, the major challenge of system-wise anomaly detection is that the activation $\boldsymbol{a}$ for the interaction with the systems is not readily available.
We have experiment results in Section~\ref{sec:exp} to illustrate that the data collected from each system based on random generated activation is representative enough for the detection of neither anomalous reward nor transition systems.
Therefore, in this work, we propose the algorithm to generate the activation so that we can collect representative reward sequence or trajectory for the detection of anomalous system.

\definecolor{gray}{RGB}{150, 150, 150}
\begin{algorithm*}
\caption{\Algnameabbr{} (\Algname{}).} % for anomalous reward detection (\Algnameabbr{}-R)}
\label{ag:alg_R}
\textbf{Inputs}: Systems $\mathcal{M}_1, \mathcal{M}_2, \cdots, \mathcal{M}_N$, \emph{anomaly\_type}. \\
\textbf{Outputs}: Policy $\mu(s \mid \theta_\mu)$ and encoder-decoder $f_D (f_E (\boldsymbol{\tau} \mid \theta_E) \mid \theta_D)$.

\begin{algorithmic}[1]

\State Randomly initialize the parameters of policy $\theta_{\mu}$ and encoder-decoder $\theta_{E}, \theta_{D}$. %, copy $\theta_{\mu}$ to the target policy $\theta_{\mu}' \gets \theta_{\mu}$.

% \State Initialize the parameters of target policy $\theta_{\mu}' \gets \theta_{\mu}$.

% \State Initialize volume relpay buffer.

\While{\emph{not converged}}

\textcolor{gray}{//--------------------------------Interact with the  systems--------------------------------//}

% \State Randomly choose a system $\mathcal{M}_i \in \{\mathcal{M}_i | 1 \leq i \leq N\}$.
\State Randomly choose a mini-batch of systems $\{ \mathcal{M}_i \}_{1 \leq i \leq B} \subseteq \{\mathcal{M}_i | 1 \leq i \leq N\}$.

% \State Interact with $\mathcal{M}_i$ based on the target policy $\mu(s | \theta_\mu')$ to generate the tuple of trajectory and reward $(\boldsymbol{\tau}_i, \boldsymbol{r}_i)$.
\State 
Interact with $\{ \mathcal{M}_i \}_{1 \leq i \leq B}$ based on the policy $\mu(s | \theta_\mu)$ to collect the tuples of trajectory and reward $(\boldsymbol{\tau}_i, \boldsymbol{r}_i)_{1 \leq i \leq B}$.

\textcolor{gray}{//--------------------------------------------System Embedding Neutralization----------------------------------------------//}

% \State Randomly choose a mini-batch of MDP systems $\{ \mathcal{M}_i \}_{1 \leq i \leq B} \subseteq \{\mathcal{M}_i | 1 \leq i \leq N\}$.

% \State Interact with $\{ \mathcal{M}_i \}_{1 \leq i \leq B}$ based on the policy $\mu(s | \theta_\mu)$ to collect the tuples of trajectory and reward $(\boldsymbol{\tau}_i, \boldsymbol{r}_i)_{1 \leq i \leq B}$.

\State Achieve the embedding of system $\mathcal{M}_i$ by $\boldsymbol{u}_i = f_E (\boldsymbol{\tau}_i \mid \theta_E)$, and the mean embedding by $\boldsymbol{u}_c = \frac{1}{B} \sum_{i=1}^B \boldsymbol{u}_i$.

\State Update the parameters of policy $\theta_\mu$ to minimize
$L_{\mu} = \sum_{i=1}^B || \boldsymbol{u}_i - \boldsymbol{u}_c ||_2$.

\textcolor{gray}{//------------------------------------------Update the encoder-decoder------------------------------------------//}

% \State Store tuple $(\boldsymbol{\tau}_i, \boldsymbol{r}_i)$ into replay buffer, and sample $B$ tuples $(\boldsymbol{\tau}'_i, \boldsymbol{r}'_i)_{1 \leq i \leq B}$ from replay buffer. \textcolor{gray}{// experience replay}
\State Store tuple $(\boldsymbol{\tau}_i, \boldsymbol{r}_i)_{1 \leq i \leq B}$ into replay buffer, and sample $B$ tuples $(\boldsymbol{\tau}'_i, \boldsymbol{r}'_i)_{1 \leq i \leq B}$ from replay buffer. \textcolor{gray}{// experience replay}

% \State Update the parameters of encoder-decoder $\theta_E$ and $\theta_D$ to minimize 
% \begin{equation}
% \setlength\abovedisplayskip{0pt}
% \setlength\belowdisplayskip{0pt}
% L_f = \sum_{i=1}^B || f_D(f_E (\boldsymbol{\tau}'_i \mid \theta_E) \mid \theta_D) - \lambda \boldsymbol{r}'_i - (1-\lambda) \boldsymbol{\tau}'_i ||_2^2,
% \nonumber
% \end{equation}
% \Statex \quad\  where $\lambda=1$ for InterSAD-R and $\lambda=0$ for InterSAD-T.

\If {\emph{anomaly\_type} \textbf{is} 
\emph{"transition"}}
\State Update the parameters of encoder-decoder $\theta_E$ and $\theta_D$ to minimize $L_f = \sum_{i=1}^B || f_D(f_E (\boldsymbol{\tau}'_i \mid \theta_E) \mid \theta_D) - \boldsymbol{\tau}'_i ||_2^2$.

\ElsIf {\emph{anomaly\_type} \textbf{is}
\emph{"reward"}}
\State Update the parameters of encoder-decoder $\theta_E$ and $\theta_D$ to minimize $L_f = \sum_{i=1}^B || f_D(f_E (\boldsymbol{\tau}'_i \mid \theta_E) \mid \theta_D) - \boldsymbol{r}'_i ||_2^2$.

\EndIf

% , where $\hat{\boldsymbol{r}}'_i = f_D(f_E (\boldsymbol{\tau}'_i \mid \theta_E) \mid \theta_D)$ $\boldsymbol{u}'_i = f (\boldsymbol{\tau}'_i \mid \theta_E)$.

% \State Update the target policy $\theta_\mu'$ by $\theta_\mu' \gets \rho \theta_\mu + (1-\rho) \theta_\mu'$ \textcolor{gray}{// soft target update}

\EndWhile
\end{algorithmic}
\end{algorithm*}

\begin{algorithm}
\caption{Interact with a system.}
\label{ag:sample_trajectory}
\textbf{Inputs}: System $\mathcal{M}_i$ and target Policy $\mu(s \mid \theta_{\mu})$ \\
\textbf{Outputs}: Tuple of trajectory and reward sequence $(\boldsymbol{\tau}_i, \boldsymbol{r}_i)$. 
\begin{algorithmic}[1]
\State $\boldsymbol{\tau}_i \gets \{ \}$, $\boldsymbol{r}_i \gets \{ \}$
\While{$t < T$}
\State Select action $a_{t} \gets \mu(s_{t} | \theta_\mu)$.
\State Observe $s_{t+1}$ and $r_t$ from MDP $\mathcal{M}_i$.
\State $\boldsymbol{\tau}_i \gets \boldsymbol{\tau}_i \cup \{ s_{t}, a_{t} \}$, $\boldsymbol{r}_i \gets \boldsymbol{r}_i \cup \{ r_t \}$.
\State $s_{t} \gets s_{t+1}$
\EndWhile
\State Flatten $\boldsymbol{\tau}_i$ and $\boldsymbol{r}_i$ into vectors.
\end{algorithmic}
\end{algorithm}

\section{\Algname{}}

% \added{[Just high-level ideas and intuitions. Such as the model has several components. Each component is for what.]} 

% An overview of \Algname{} (\Algnameabbr{}) is shown in Figure~\ref{fig:SEN_block}.
% The key idea of \Algnameabbr{} is to encourage normal systems to generate consistent trajectories, which is motivated by one-class classification~\cite{ruff2018deep}.

We propose \Algnameabbr{} for the detection of anomalous systems.
An overview of \Algnameabbr{} is shown in Figure~\ref{fig:SEN_block}. 
First, we train a policy $\mu(s \mid \theta_{\mu})$ to generate activation for the interaction with the systems, where the activation to system $\mathcal{M}_i$ in state $s_{i,t}$ is generated by $a_{i,t} = \mu(s_{i,t} \mid \theta_{\mu})$ for $0 \leq t \leq T$ and $1 \leq i \leq N$.
Then, we design an encoder-decoder based model $f=f_D(f_E(\boldsymbol{\tau} \mid \theta_{E}) \mid \theta_{D})$ to learn embeddings for systems with their trajectories observed. Such a model structure also enables the end-to-end training of the policy network. The system embeddings and reward signals are used in measuring anomaly scores. Meanwhile, to stabilize the training process, \Algnameabbr{} uses Experience Replay Mechanism to collect more stable data batches. The details of our approach are introduced in the subsections below.

\subsection{The Encoder-Decoder Models}
\label{sec:31}

The encoder-decoder model $f_D(f_E(\boldsymbol{\tau}_i \mid \theta_{E}) \mid \theta_{D})$ uses the $f_E$ module to learn the embedding of each system $\mathcal{M}_i$ from its generated trajectory $\boldsymbol{\tau}_i$.
Since anomaly detection uses reward sequences and state transitions to detect anomalous reward and transition systems, respectively, we propose \Algnameabbr{}-T and \Algnameabbr{}-R which employ different supervised signal collected from each system to train the model.

\Algnameabbr{}-T works for the detection of anomalous transition systems that have different state transition compared with normal systems.
Specifically, the encoder $f_E: \mathcal{S}^T \!\times\! \mathcal{A}^T \to \mathbb{R}^{D}$ learns the system embedding from trajectory, and the decoder $f_D: \mathbb{R}^{D} \to \mathcal{S}^T \!\times\! \mathcal{A}^T$ receives the embedding to reconstruct the trajectory, so that the embedding reserves transition-relevant characteristics of system $\mathcal{M}_i$, where $D$ is the dimension of embedding space.
Specifically, the encoding-decoding process is written as:
\begin{equation}
\begin{aligned}
\boldsymbol{u}_i &= f_E ( \boldsymbol{\tau}_i \mid \theta_E),
\\
\hat{\boldsymbol{\tau}}_i &= f_D ( \boldsymbol{u}_i \mid \theta_D),
\end{aligned}
\end{equation}
where $\boldsymbol{u}_i$ is the embedding of $\mathcal{M}_i$, and $\hat{\boldsymbol{\tau}}_i$ denotes the reconstructed trajectory. The model is trained to minimize the reconstruction error, so $\theta_{E}^*, \theta_{D}^* = \arg \min \sum_{i=1}^{N} ||\boldsymbol{\tau}_i - \hat{\boldsymbol{\tau}}_i||^2_2$.

\Algnameabbr{}-R works for detecting anomalous reward systems that have different distributions of reward signal from normal systems.
\Algnameabbr{}-R has the same encoder $f_E: \mathcal{S}^T \!\times\! \mathcal{A}^T \to \mathbb{R}^{D}$ as \Algnameabbr{}-T to learn the system embedding.
However, \Algnameabbr{}-R feeds the embedding to the decoder $f_D: \mathbb{R}^{D} \to \mathbb{R}^T$ to approximate the reward sequence $\boldsymbol{r}_i$.
\Algnameabbr{}-R uses the reward sequence $\boldsymbol{r}_i$ as the supervised signal to update the model so that the embedding $\boldsymbol{u}_i \in \mathbb{R}^{D}$ contains reward-relevant characteristics of system $\mathcal{M}_i$.
Specifically, \Algnameabbr{}-R has encoding-decoding process given by:
\begin{equation}
\begin{aligned}
\boldsymbol{u}_i &= f_E ( \boldsymbol{\tau}_i \mid \theta_E),
\\
\hat{\boldsymbol{r}}_i &= f_D ( \boldsymbol{u}_i \mid \theta_D).
\end{aligned}
\end{equation}
The model is trained to minimize the error, so $\theta_{E}^*, \theta_{D}^* = \arg \min \sum_{i=1}^{N} ||\boldsymbol{r}_i - \hat{\boldsymbol{r}}_i||^2_2$.

% Daochen The goal of the embedding network is to map the the generated trajectory or reward sequences into embedding space.

\begin{figure}
\centering
\includegraphics[width=0.4\textwidth]{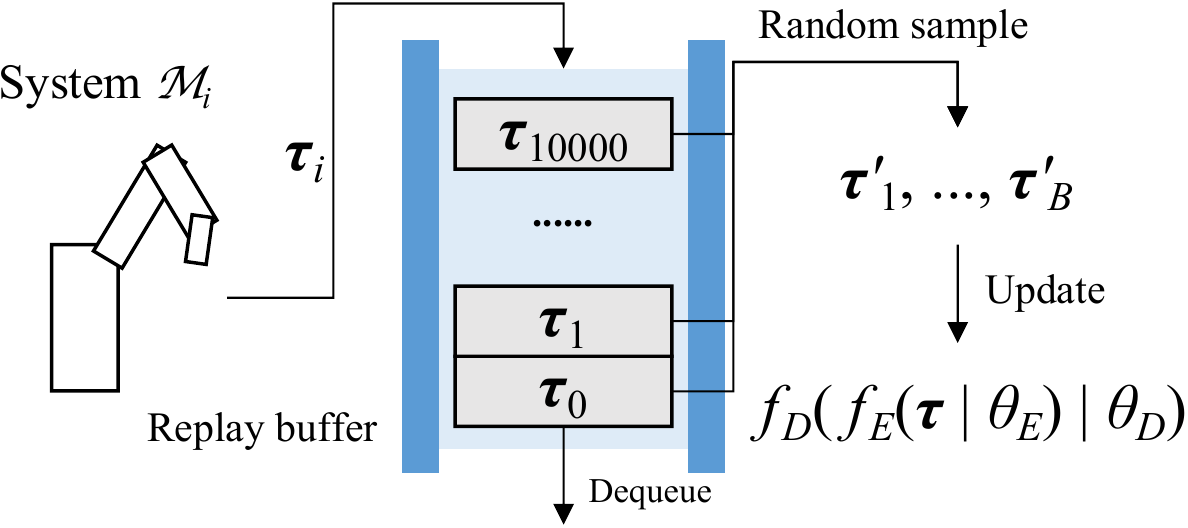}
\caption{Experience replay mechanism (ERM).}
\label{fig:ERM}
\end{figure}

% \vspace{-2mm}
\subsection{System Embedding Neutralization}
\label{sec:32}

% encourages normal MDP systems to generate consistent rewards or trajectories across different activation trials to reach an equilibrium that is less susceptible to randomness.

\Algnameabbr{} learns the policy $\mu(s | \theta_{\mu})$ as a neural network to provide activation for the interaction with the systems.
Specifically, the policy generates the activation $a_{i,t} = \mu(s_{i,t} | \theta_{\mu})$ for system $\mathcal{M}_i$ and collect $s_{i,t+1}, r_{i,t}$ iteratively for $0 \leq t \leq T$, so that the trajectories $\{ \boldsymbol{\tau}_i \}_{1 \leq i \leq N}$ and rewards $\{ \boldsymbol{r}_i \}_{1 \leq i \leq N}$ can be collected for anomaly detection.
However, the response generated from each system can significantly vary given different activations, which makes the anomalous systems less distinguishable from the normal ones, thus increasing the likelihood of false alarm and miss detection. Therefore, \Algnameabbr{} updates the policy $\mu(s \mid \theta_{\mu})$ to \textit{neutralize} the system embeddings $\{ \boldsymbol{u}_i \}_{1 \leq i \leq N}$, where the embedding of system $\mathcal{M}_i$ is given by $\boldsymbol{u}_i = f_E(\boldsymbol{\tau}_i \mid \theta_E)$.
Specifically, we propose System embedding neutralization (SEN) to minimize the overall pairwise distance between the embeddings of the $N$ systems:
\begin{equation}
\begin{aligned}
\label{eq:optimal_mu1}
\theta_\mu^* = \arg\min \sum_{i=1}^N \sum_{j=1}^N || \boldsymbol{u}_i - \boldsymbol{u}_j ||_2,
\end{aligned}
\end{equation}
where $\boldsymbol{u}_i = f_E(\boldsymbol{\tau}_i \mid \theta_E)$, $\boldsymbol{\tau}_i \!\!=\! [s_{i,0}, a_{i,0}, \cdots, s_{i,T-1}, a_{i,T-1}]$, and $a_{i,t} = \mu(s_{i,t} \mid \theta_{\mu})$ for $0 \leq t \leq T$.
% The objective of SEN is to encourage the interaction with each system to be less susceptible to randomness.
The numerical results in Section~\ref{sec:exp} indicate that the performance of anomaly detection is inversely correlated with the overall variance of system embeddings, and demonstrate that SEN can reduce the variance during the training process as well as improve the performance of system-wise anomaly detection.

% \footnote{The target sequence is denoted as the reward sequence $\boldsymbol{r} = [r_0, \cdots, r_{T-1}]$ or trajectory $\boldsymbol{\tau} = [s_0, a_0, \cdots, s_{T-1}, a_{T-1}]$ in detecting anomalous reward or transitions MDP systems, respectively.} 
% from normal systems as possible, where tiny distance between the target sequences leads to low score for normal systems.

However, it is inefficient to directly update the policy network based on Equation~(\ref{eq:optimal_mu1}), where the computational complexity is $O(N^2)$.
To improve efficiency, we plug $|| \boldsymbol{u}_i - \boldsymbol{u}_j ||_2 \leq || \boldsymbol{u}_i - \boldsymbol{u}_c ||_2 + || \boldsymbol{u}_j - \boldsymbol{u}_c ||_2$ into Equation~(\ref{eq:optimal_mu1}) and get
$\sum_{i=1}^N \sum_{j=1}^N || \boldsymbol{u}_i - \boldsymbol{u}_j ||_2 \leq 2(N-1) \sum_{i=1}^N || \boldsymbol{u}_i - \boldsymbol{u}_c ||_2$, where $\boldsymbol{u}_c = \frac{1}{N} \sum_{i=1}^{N} \boldsymbol{u}_i$ denotes the mean value of embeddings.
Finally, \Algnameabbr{} updates the policy to minimize the upper bound of Equation~(\ref{eq:optimal_mu1}) given by:
\begin{equation}
\begin{aligned}
\label{eq:optimal_mu2}
\theta_\mu^* = \arg\min \sum_{i=1}^N || \boldsymbol{u}_i - \boldsymbol{u}_c ||_2,
\end{aligned}
\end{equation}
where the computational complexity is $O(N)$, and the relevant proof is given in \emph{Appendix A}. This objective encourages $\boldsymbol{u}_i$ for $1 \leq i \leq N$ to approach its mean value $\boldsymbol{u}_c$, and thus achieves neutralization of system embeddings.

\subsection{Experience Replay Mechanism}

The data to train our encoder-decoder model $f$ are obtained from the real-time interactions with systems. However, the distribution of the real-time collected data from each system is non-stationary and noisy, which leads to an unstable and even divergent training process.
To solve this problem, we adopt Experience Replay Mechanism (ERM)~\cite{lin1992self,zha2019experience} to update the encoder-decoder model in an offline manner.
To be more concrete, as shown in Figure~\ref{fig:ERM}, after each interaction with the systems, the latest tuples of trajectory and reward sequence are enqueued to a \textit{replay buffer} waiting to be sampled, and the earliest tuples will be dequeued once the size of the buffer reaches a preset limitation, which is set to $10^6$ in our experiments. 
The model $f$ is not updated using the latest trajectory or reward signal, but using the tuples sampled from the replay buffer.
With mini-batch updating, we randomly sampled $B$ tuples from the replay buffer to update $f$ in each iteration, where $B$ denotes the mini-batch size.
Experience replay mechanism enables $f$ to be updated based on the mixture of samples generated in different time stages from the replay buffer. It provides a more smoothed transition of the distribution of data for training the model, thus leading to stable learning and faster convergence.
We have empirically studied the effectiveness of ERM through experiments reported in Section~\ref{exp:5}.

% generated from the interaction with each system.
% Detailed description and interpretation on soft target update and experience replay mechanism can be referred to  and , respectively.

\subsection{Summary of Training \Algnameabbr{}}

The training of \Algnameabbr{} is summarized in Algorithm~\ref{ag:alg_R}. 
Specifically, both \Algnameabbr{}-T and \Algnameabbr{}-R have the same high-level structure: 
(i) begins with random initializing the policy and encoder-decoder (line~1).
(ii) execute the iterative process including the interaction with the systems by Algorithm~\ref{ag:sample_trajectory} (lines 3-4), the update of the policy by SEN (lines~5-6) and the update of encoder and decoder by ERM (lines~7-11).
(iii) comes to an end with the convergence of the policy, encoder and decoder (line~2).

\Algnameabbr{}-T updates the decoder in different ways from \Algnameabbr{}-R, as a result of targeting the detection of different type of anomalous systems.
For the detection of anomalous transition system, \Algnameabbr{}-T learns the system embedding which can reserve the transition-relevant characteristic, thus minimizing the reconstruction error of using the system embedding to reconstruct the collected trajectory~(line~9);
for the detection of anomalous reward system \Algnameabbr{}-R minimize the MSE error supervised by the collected rewards in order that the reward-relevant characteristic can be reserved~(line~11).
\subsection{Anomaly Score Estimation}
\label{sec:33}

\Algnameabbr{} outputs the trained policy $\mu(s \mid \theta_{\mu})$ to generate activation for interacting with the systems, obtaining system responses as features, and detecting their abnormal characteristics. Specifically, the policy interacts with each system $\mathcal{M}_i$ to obtain $\boldsymbol{\tau}_i$ and $\boldsymbol{r}_i$.
Then, the anomaly score of $\mathcal{M}_i$ is measured by $\phi(\boldsymbol{\tau}_i; \{ \boldsymbol{\tau}_j \}_{1 \leq j \leq N})$ and $\phi(\boldsymbol{r}_i, \{ \boldsymbol{r}_j \}_{1 \leq j \leq N})$ in the detection of anomalous transition and anomalous reward systems, respectively. 
Here $\phi(\cdot ,  \cdot)$ can be referred to a certain anomaly detector where features of instances are given as input~\cite{kriegel2008angle, Liu2012, Breunig2000, Manevitz2002}.
Meanwhile, in this work, we also investigate using system embeddings as an alternative option for detecting anomalous transition systems, and conduct the validating experiment in Section~\ref{exp:4}.
In this case, the trajectories $\{ \boldsymbol{\tau}_i \}_{1 \leq i \leq N}$ are fed into the encoder to get embeddings $\boldsymbol{u}_i = f_E(\boldsymbol{\tau}_i | \theta_E)$, where $\boldsymbol{u}_i$ reserves transition-relevant characteristics for the system in \Algnameabbr{}-T, so that the anomaly score is $\phi(\boldsymbol{u}_i; \{ \boldsymbol{u}_j \}_{1 \leq j \leq N})$ for system $\mathcal{M}_i$\footnote{However, the anomaly score for detecting anomalous rewards cannot be $\phi(\boldsymbol{u}_i, \{ \boldsymbol{u}_j \}_{1 \leq j \leq N})$ even though $\boldsymbol{u}_i$ reserves reward-relevant component in \Algnameabbr{}-R, since $\boldsymbol{u}_i$ derives from $\boldsymbol{\tau}_i$, which is not out-of-distribution for anomalous reward systems.}.

% Then, any existing anomaly detection algorithm can be applied to estimate anomaly score for each system based on its representation, such as OCSVM, LOF or $i$Forest.
% $\text{Afterwards}$, those attached with top scores are anomalous systems.

% The estimation of anomaly score for each system based on its representation can be referred to any existing anomaly detection algorithm,} such as OCSVM~\cite{Manevitz2002}, LOF~\cite{Breunig2000} or $i$Forest~\cite{Liu2012}.

% algorithm to estimate the anomaly score for each system based on its representation can be refered to existing work, such as OCSVM~\cite{Manevitz2002}, LOF~\cite{Breunig2000} or $i$Forest~\cite{Liu2012}.

% reward sequences $\boldsymbol{r}_i$ and trajectories $\boldsymbol{\tau}_i$ for $1 \leq i \leq N$ can be collected from the $K$ MDP systems.
% For each system $\mathcal{M}_i$, its anomaly score is estimated based on the reward sequence $\boldsymbol{r}_i$ as its representation in the detection of anomalous reward systems, and based on the embedded trajectory $\boldsymbol{u}_i = f (\boldsymbol{\tau}_i \mid \theta_f)$ in the detection of anomalous transition systems.

\begin{figure*}
% \setlength{\abovecaptionskip}{-1mm}
% \setlength{\belowcaptionskip}{-5mm}
% \subfigcapskip=-1mm
\centering
\begin{minipage}{0.32\linewidth}
\centering
\subfigure[]{
\centering
\includegraphics[width=1.0\textwidth]{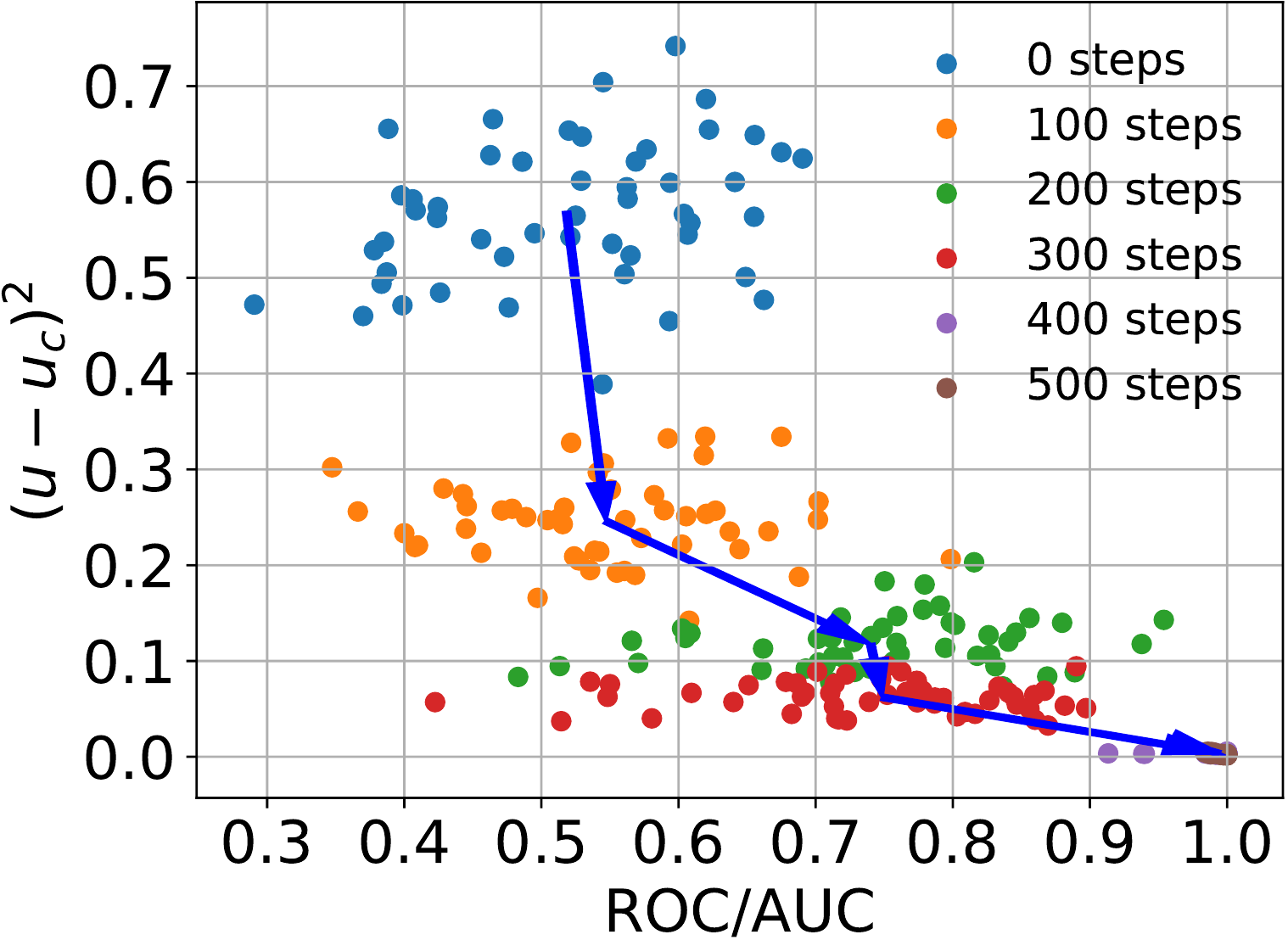}
}
\end{minipage}
\text{ }
\begin{minipage}{0.32\linewidth}
\centering
\subfigure[]{
\centering
\includegraphics[width=1.0\textwidth]{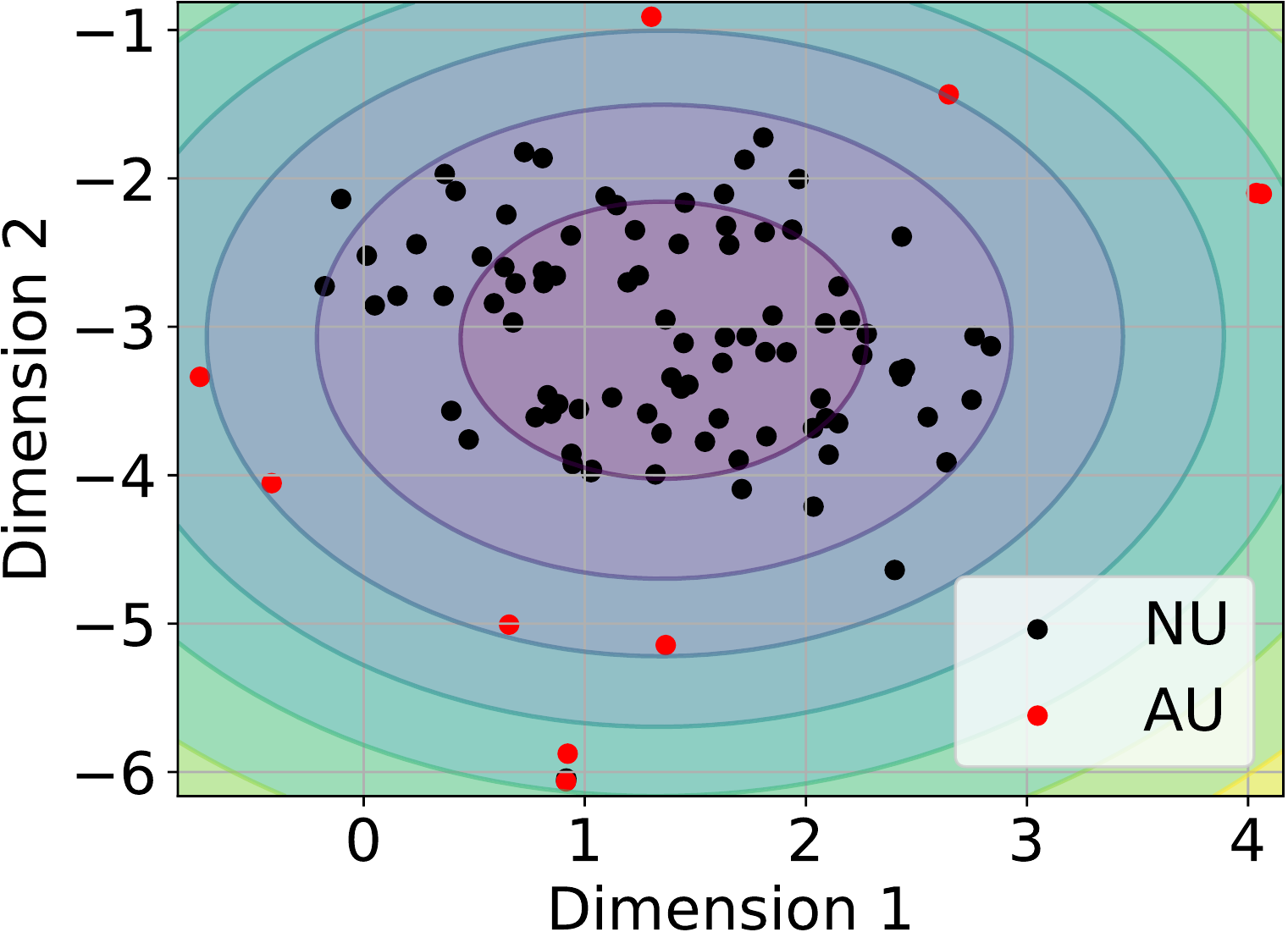}
}
\end{minipage}
\text{ }
\begin{minipage}{0.32\linewidth}
\centering
\subfigure[]{
\centering
\includegraphics[width=1.0\textwidth]{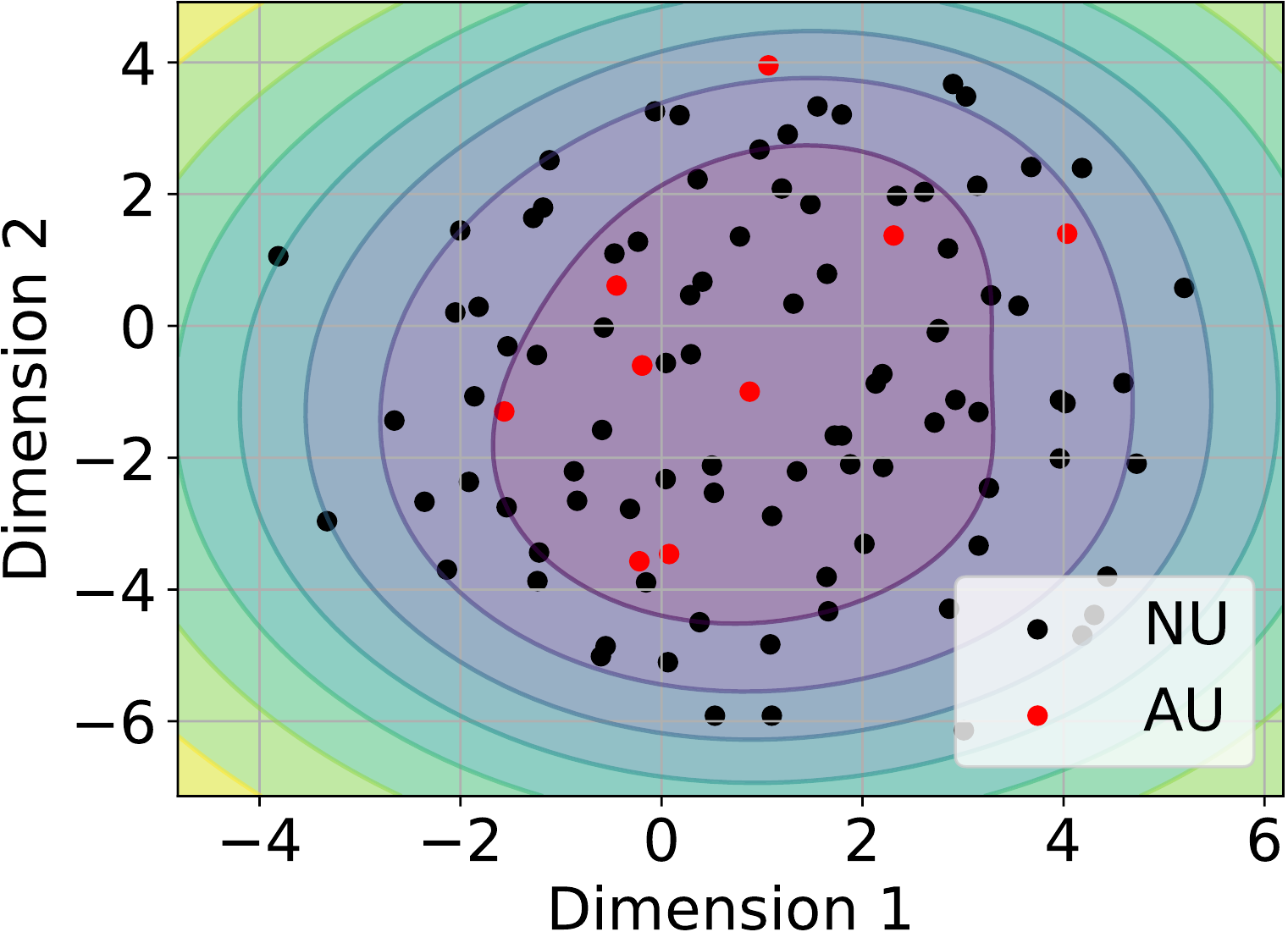}
}
\end{minipage}
\caption{(a) 50 independent testings of the policy trained for $0, 100, \cdots, 500$ iterations; X label: ROC-AUC of anomaly detection; Y label: $\sum_{i=1}^N||\boldsymbol{u}_i - \boldsymbol{u}_c||^2_2$; Arrow: Tracking the training of \Algnameabbr{}-R. 
(b) and (c) 2D embedding of the reward sequence collected from the systems during the interactions with \Algnameabbr{}-R (b) and random recommendation (c); NU and AU: the rewards collected from normal and anomalous users, respectively.
}
\label{fig:virtual_taobao_2D_embedding}
% \begin{minipage}{0.33\linewidth}
% \centering
% \subfigure[]{
% \centering
% \includegraphics[width=1.0\textwidth]{figure/robotic_interaction-cropped.pdf}
% }
% \end{minipage}
% \text{    }
% \begin{minipage}{0.2\linewidth}
% \centering
% \subfigure[]{
% \centering
% \includegraphics[width=1.0\textwidth]{figure/mujoco.png}
% }
% \end{minipage}
\end{figure*}

\section{Experiments}
\label{sec:exp}

%We evaluate \Algnameabbr{} upon real-world tasks, including the detection of user attack in recommender system and identifying anomalous robotic systems.
%Specifically, the experiments are conducted to answer the following three questions:
We conduct experiments to answer the following research questions.
\textbf{RQ1}: How does \Algnameabbr{} perform compared with baselines in the detection of anomalous transition and anomalous reward system, respectively~(Sec.~\ref{exp:1.1} and \ref{exp:1.2})?
\textbf{RQ2}: How does SEN contribute to the detection of anomalous system~(Sec.~\ref{exp:2})?
\textbf{RQ3}: Are SEN and ERM both necessary for \Algnameabbr{}~(Sec.~\ref{exp:3})?
\textbf{RQ4}: Can we use the system embedding to estimate anomaly score? (Sec.~\ref{exp:4})?
\textbf{RQ5}: How will the hyperparameters impact the performance of \Algnameabbr{}~(Sec.~\ref{exp:5})?

% \textbf{RQ2}: How does \Algnameabbr{}-R perform compared with baselines in the detection of anomalous reward system~(Sec.~\ref{exp:1})?
% \textbf{RQ4}: How to intuitively understand the isolation of systems in embedding space~(Sec.~\ref{exp:3})?

\begin{table} % [htp]\small
\centering
% \begin{minipage}[t]{0.5\textwidth}
\centering
\caption{Hyperparameters of \Algnameabbr{}-T.}
\vspace{-2mm}
\begin{tabular}{l|c|c|c}
  \hline
  Hyperparameter & HalfCheetah & Walker2D & Ant \\ 
  \hline
  State space $\mathcal{S}$ & $\mathbb{R}^{26}$ & $\mathbb{R}^{22}$ & $\mathbb{R}^{28}$ \\
  \hline 
  Activation space $\mathcal{A}$ & $\mathbb{R}^{6}$ & $\mathbb{R}^{6}$ & $\mathbb{R}^{8}$ \\
  \hline
  Embedding dim $D$ & 160 & 140 & 180 \\
  \hline 
  LSTM hidden dim & \multicolumn{2}{c|}{$32$} & $64$ \\
  \hline
  Max time step $T$ & \multicolumn{3}{c}{$10$} \\
%   \hline
  Learning rate & \multicolumn{3}{c}{$0.001$} \\
%   \hline
  Mini-batch size $B$ & \multicolumn{3}{c}{$32$} \\
%   \hline
  Replay buffer size & \multicolumn{3}{c}{$10^6$} \\
  \hline
  Observation noise & $0.02$ & $0.05$ & $0.04$ \\
  \hline
  Training interations & 4500 & 4000 & 1000 \\
  \hline
\end{tabular}
\label{tb:hyperparameters_SEN/T}
% \end{minipage}
\end{table}

% Specifically, for each difficulty level $\lambda = 10, 20$ and $30$, \Algnameabbr{} shows highest ROC-AUC.
% Furthermore, it is convincing that all methods degrade as the difficulty level grows, given by $\lambda$ reducing from $30\%$ to $10\%$.

\subsection{Anomalous Transition Systems Detection (RQ1)}
\label{exp:1.1}

In this experiment, we use robotic interaction as the scenario to evaluate the performance of \Algnameabbr{}-T in identifying anomalous transition systems.
For a robotic system $\mathcal{M}_i$, at each time point $t$, the state $s_{i,t}$ includes a robot's position, velocity, and angles of joints; action $a_{i,t}$ is the external electrical activation to the system; and the state transition $s_{i,t} \to s_{i,t+1}$ characterize a robot's reaction to the activation~\cite{brockman2016openai}.
% Anomalous robot have out-of common reactions in contrast normal ones.

% \vspace{-1mm}
\subsubsection{Experimental Settings}
\begin{figure*}
% \setlength{\abovecaptionskip}{-1mm}
% \setlength{\belowcaptionskip}{-5mm}
% \subfigcapskip=-1mm
\centering
\begin{minipage}{0.25\linewidth}
\centering
\subfigure[\emph{OpenAI/Gym/Pybullet} Preview.]{
\centering
\includegraphics[width=1.0\textwidth]{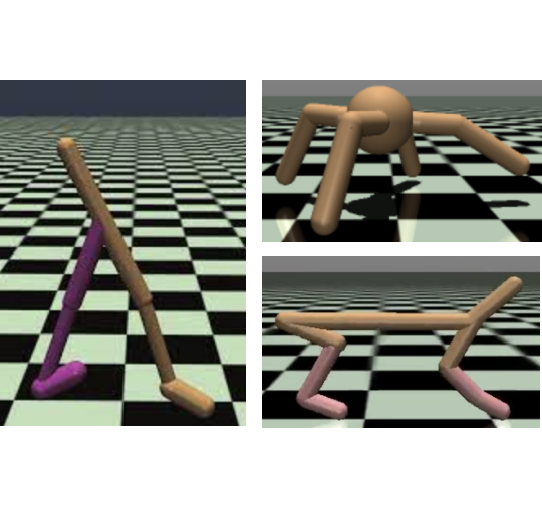}}
\end{minipage}
\text{ }
\text{ }
\begin{minipage}{0.34\linewidth}
\centering
\subfigure[Detection of A-\emph{HalfCheetah}.]{
\centering
\includegraphics[width=1.0\textwidth]{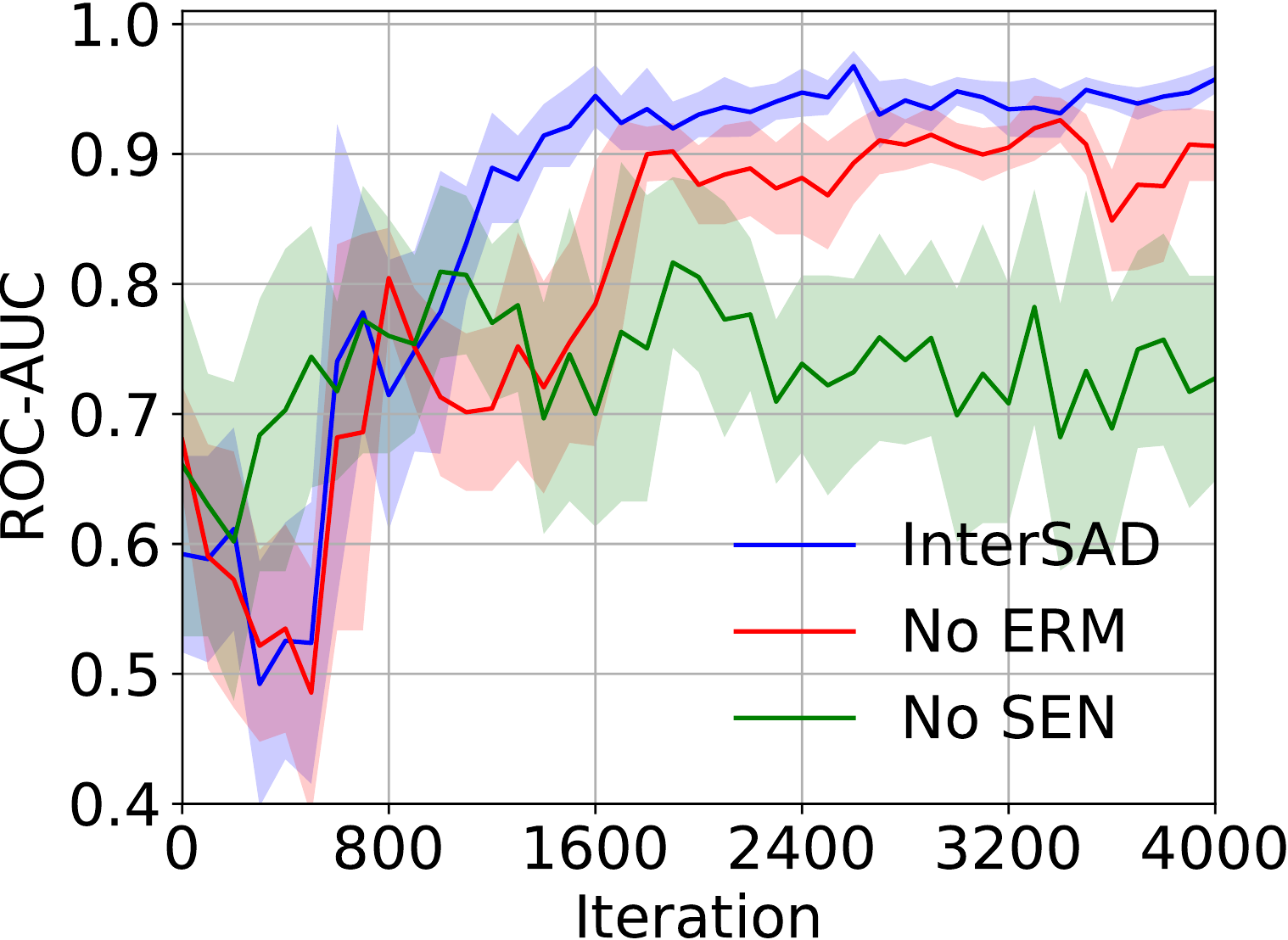}
}
\end{minipage}
\begin{minipage}{0.34\linewidth}
\centering
\subfigure[Detection of A-\emph{Walker2D}.]{
\centering
\includegraphics[width=1.0\textwidth]{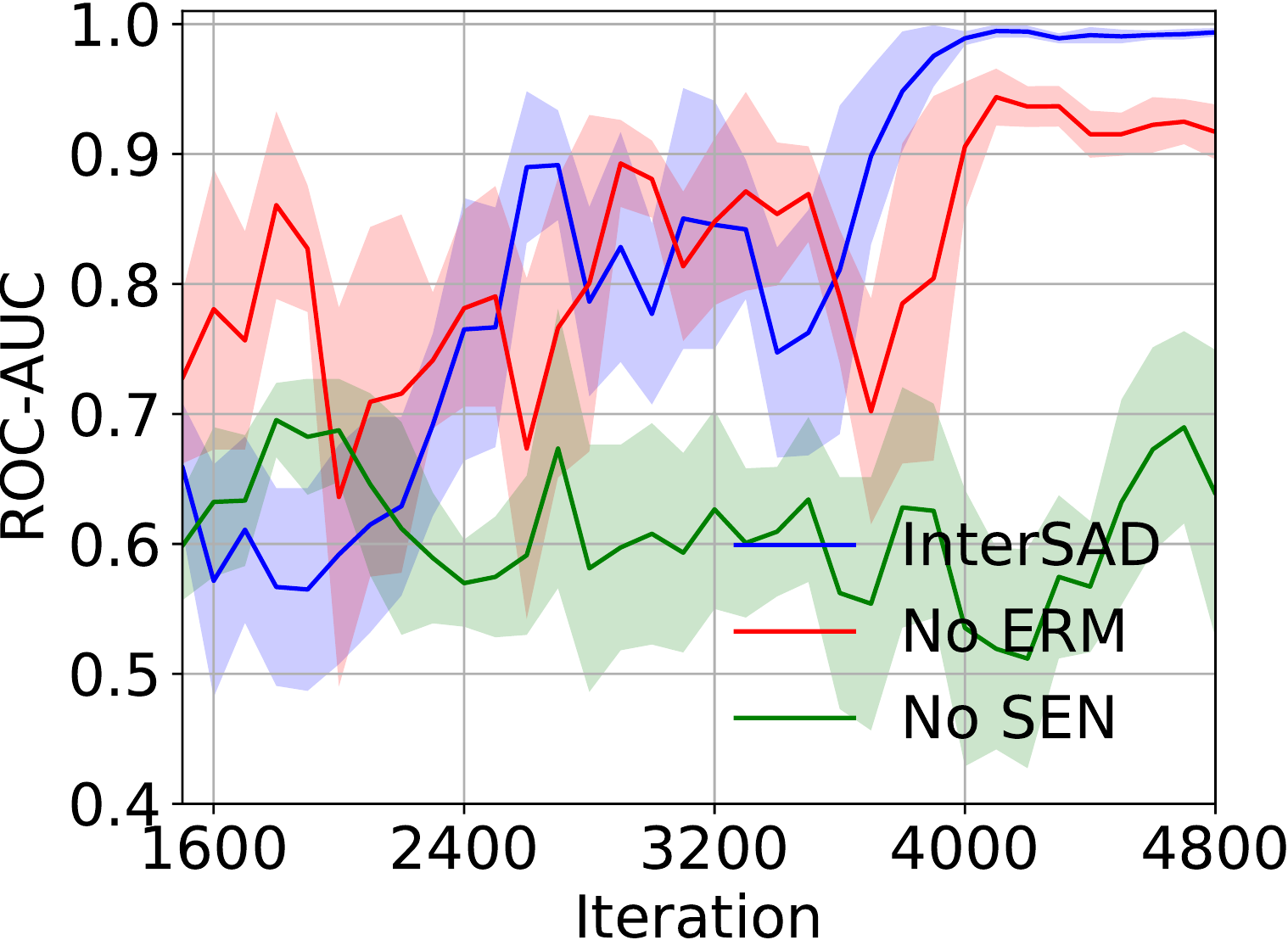}
}
\end{minipage}
\caption{(a) Preview of robots Walker2D, Ant and HalfCheetah from \emph{OpenAI/Gym/Pybullet}.
(b)-(c) Performance of \Algnameabbr{}-T and that without SEN or ERM.}
\label{fig:ablation_results}
\end{figure*}

The experiments are based on OpenAI/Gym/Pybullet\footnote{\url{https://github.com/bulletphysics/bullet3}}, which provides APIs to simulate different types of robotic systems including \emph{HalfCheetah}, \emph{Walker2D} and \emph{Ant} shown in Figure \ref{fig:ablation_results} (a).
\Algnameabbr{}-T is trained using the hyperparameters given in Table~\ref{tb:hyperparameters_SEN/T}, and is tested based on
the anomalous systems generated by OpenAI/Gym/Pybullet including anomalous \emph{HalfCheetah} (A-\emph{HalfCheetah}), \emph{Walker2D} (A-\emph{Walker2D}) and \emph{Ant} (A-\emph{Ant}).
In our experiments, anomalies are initialized with abnormal physical parameters, which indirectly lead to state transitions different from others.
Specifically, A-\emph{HalfCheetah} has $50$mm diameters of the thigh, shin, and foot in contrast with normal $46$mm;
A-\emph{Walker2D} has \{$55, 45, 65$\}mm diameters of thighs, shins and feet, respectively, in contrast with normal \{$50, 40, 60$\}mm;
A-\emph{Ant} has $12 \!\times\! 88$mm diameters of legs in contrast with normal $12 \!\times\! 80$mm. % more details are provided in Appendix C.
After learning with 2000 iterations, \Algnameabbr{}-T is tested on 1000 systems including 90$\%$ normal and 10$\%$ anomalous systems, where the maximum time step of the interaction with the systems $T=10$.
%The detectors include Deep Deterministic Policy Gradient (DDPG) and Random Sampling with $i$Forest (RS-$i$Forest), LOF (RS-LOF), and OCSVM (RS-OCSVM) in the same conditions with \Algnameabbr{}-T as baselines for comparison, which are specified as follows:
The baseline methods are introduces as below.

\begin{itemize}[leftmargin=11pt]
\setlength{\parskip}{-3.5pt}
\setlength{\itemsep}{5pt}
\setlength{\partopsep}{0pt}
\item 
\textbf{Deep Deterministic Policy Gradient (DDPG)}:
We use DDPG~\cite{lillicrap2015continuous} to learn the policy for the interaction with each system to collect the trajectories $\{ \boldsymbol{\tau}_i \}_{1 \leq i \leq N}$.
The anomaly score $\phi( \boldsymbol{\tau}_i; \{ \boldsymbol{\tau}_i \}_{1 \leq i \leq N} )$ of each system is estimated using $i$Forest~\cite{Liu2012}.

\item 
\textbf{Random Sampling Based Detectors}: Besides using carefully-designed activations, we also generate random activations to collect the trajectories $\{ \boldsymbol{\tau}_i \}_{1 \leq i \leq N}$ from the systems.
The anomalous systems are indicated by the anomaly score is computed using $i$Forest~\cite{Liu2012}, LOF~\cite{Breunig2000} and OCSVM~\cite{Manevitz2002}. These detectors are called RS-$i$Forest, RS-LOF, and RS-OCSVM.
\end{itemize}
% \vspace{-1mm}
\noindent
% In addition, Additive Gaussian noise is introduced for both normal systems and anomalies to simulate the observation noise in the real world. % Standard deviation
% The performance of anomaly detection is evaluated by ROC-AUC. The experiment results are reported in Table~\ref{tb:gym_exp_result}.

% Guanchu: Furthermore, additive Gaussian noise satisfying $\mathcal{N}(0, 0.05)$ is introduced for both normal systems and anomalies to simulate the observation noise in the real world. 

\subsubsection{\Algnameabbr{}-T compared with baselines}

The ROC-AUC of anomalous systems detection result is reported in Table~\ref{tb:gym_exp_result}, where \Algnameabbr{}-T consistently performs better than baseline methods in the detection of all type of anomalous transition systems.
In contrast with \Algnameabbr{}-T, the trajectories collected from random interactions with the systems fail to provide representative features to distinguish anomalies from normal systems. Meanwhile, the policy learned by DDPG provides the activation for maximizing the cumulative reward but fails to isolate anomalous systems, which leads to poor performance.

\begin{table}
% \begin{minipage}[t]{0.45\textwidth}
\centering
\caption{Physical parameters of anomalous robotic systems inconsistent with normal systems in our experiment.}
\vspace{-2mm}
\begin{tabular}{l|c|c}
  \hline
    \makecell[l]{Robotic \\ system}
    & \makecell[c]{Diameters of thigh, \\ shin and foot}
    & Diameters of leg \\
  \hline
  \emph{HalfCheetah}
   & \{46, 46, 46\}mm & \multirow{4}{*}{\diagbox[width=8em,height=4\line]{\phantom{x}}{\phantom{x}}} \\
  \cline{1-2}
  A-\emph{HalfCheetah} & \{50, 50, 50\}mm & \\
  \cline{1-2}
  \emph{Walker2D} & \{50, 40, 60\}mm & \\
  \cline{1-2}
  A-\emph{Walker2D} & \{55, 45, 65\}mm & \\
  \hline
  \emph{Ant} & \multirow{2}{*}{\diagbox[width=10em,height=2\line]{\phantom{x}}{\phantom{x}}} & 80mm \\
  \cline{1-1}
  \cline{3-3}
  A-\emph{Ant} & & 88mm \\
  \hline
\end{tabular}
\label{tb:physical_parameters}
% \end{minipage}
\end{table}

\begin{table}[] \small
    \centering
    \caption{ROC-AUC of detecting anomalous robotic systems.}
    \vspace{-2mm}
    \begin{tabular}{l|c|c|c}
      \hline
        \diagbox [width=5em,trim=l]    {Detector}{Type}
        & \!A-\emph{HalfCheetah}\!
        & A-\emph{Walker2D}
        & A-\emph{Ant} \\ % Hopper iForest $0.81$ LOF $0.79$ OCSVM $0.66$
      \hline
% =============== iForest ================
% \!\!RS-$i$Forest\!\! & 0.744 $\!\!\pm\!\!$ 0.054 & 0.786 $\!\!\pm\!\!$ 0.059 & 0.863 $\!\!\pm\!\!$ 0.114 \\
RS-$i$Forest\!\! & 0.767 $\!\!\pm\!\!$ 0.079 & 0.824 $\!\!\pm\!\!$ 0.126 & 0.812 $\!\!\pm\!\!$ 0.088 \\
% =============== OCSNM ================
% \!\!RS-OCSVM\!\! & 0.788 $\!\!\pm\!\!$ 0.166 & 0.710 $\!\!\pm\!\!$ 0.086 & 0.792 $\!\!\pm\!\!$ 0.121 \\
RS-OCSVM\!\! & 0.640 $\!\!\pm\!\!$ 0.186 & 0.720 $\!\!\pm\!\!$ 0.121 & 0.714 $\!\!\pm\!\!$ 0.064 \\
% =============== LOF ================
% RS-LOF & 0.858 $\!\!\pm\!\!$ 0.101 & 0.918 $\!\!\pm\!\!$ 0.051 & 0.952 $\!\!\pm\!\!$ 0.036 \\
RS-LOF & 0.657 $\!\!\pm\!\!$ 0.169 & 0.765 $\!\!\pm\!\!$ 0.091 & 0.801 $\!\!\pm\!\!$ 0.144 \\
% =============== DDPG ================
DDPG & 0.807 $\!\!\pm\!\!$ 0.027 & 0.751 $\!\!\pm\!\!$ 0.014 & 0.897 $\!\!\pm\!\!$ 0.0256 \\ 
% =============== DeepSAD ================
% \textbf{DeepSAD-T} &\textbf{0.987} $\!\!\pm\!\!$ 0.006&\textbf{1.0} $\!\!\pm\!\!$ 0. &\textbf{0.999} $\!\!\pm\!\!$ 0.001 \\
\textbf{\Algnameabbr{}-T} &\textbf{0.930} $\!\!\pm\!\!$ 0.030 &\textbf{0.999} $\!\!\pm\!\!$ 0.001 &\textbf{0.999} $\!\!\pm\!\!$ 0.001 \\
\hline
\end{tabular}
\label{tb:gym_exp_result}
\end{table}
% Do not remove!
% Comments: LOF turns the worst with more anomalous systems.

% generalization ability against environment's variance enables the different systems.
% enables the policy to sample a nearly optimal trajectory from anomalous systems, which fails to provide valuable features for anomaly detection.
% Different from \Algnameabbr{}, DDPG policy aims to sample an optimal trajectory from normal systems to maximize the cumulative reward, and has strong generalization ability against the divergence of different systems.

\subsection{Attack Detection in Recommender Systems (RQ1)}
\label{exp:1.2}

In this experiment, we use recommendation as the scenario to evaluate the performance of \Algnameabbr{}-R in identifying malicious users. \textit{User attack} is a popular attack model in recommender system~\cite{yang2017detecting, seminario2014attacking, wilson2013power}.
Specifically, anomalous users deliberately make fake clicks on $a \%$ of items (e.g., click undesirable items or skip favourite items), so that the system mistakenly overestimates or underestimates the item values, which may severely affect its recommendations to normal users.
In this experiment, we treat each user $i$ as a system $\mathcal{M}_i$. The user features, the items recommended to the users, and users' clicks on the items in each time point $t$ are regarded as the state $s_{i,t}$, activation $a_{i,t}$, and reward $r_{i,t}$, respectively. The interest shift of user $i$ from time point $t$ to $t+1$ is characterized by the state transition $s_{i,t} \to s_{i,t+1}$.

% \vspace{-1mm}
\subsubsection{Experimental Experimental Settings}
The experiment is based on  \emph{VirtualTaobao}\footnote{\url{https://github.com/eyounx/VirtualTaobao}}, which maps user attributes, recommended items, and the user' clicks on the items to the 91-dimensional state, 26-dimensional activation and 1-dimensional reward, respectively, and provides APIs to generate and interact with users~\cite{shi2019virtual}. % More details can be found in Appendix B.
We list the hyperparameters of \Algnameabbr{}-R in Table~\ref{tb:hyperparameters_SEN/R}. \Algnameabbr{}-R is trained for 1000 iterations and tested based on a separate set of 10000 users generated by \emph{VirtualTaobao} including 99$\%$ normal users and 1$\%$ anomalous users, where it recommends 10 times$\times$20 items/time to each user.
ROC-AUC is employed as the evaluation metric.
The baseline methods used for comparison are introduced as below.

% with OCSVM 
% In the training stage, normal customers are generated by \emph{VirtualTaobao} and are involved in the learning of policy $\mu$ and reward predictor $LSTM_R$.
% Specifically, in each time step, we use the policy $\mu$ to sample a length of $10$ trajectory from each users, and a size of $32$ mini-batch is sampled from the replay buffer and new customers to update the parameters of reward predictor $f $ and policy $\mu$, respectively.
% The hyperparameters of \Algnameabbr{}-R are given in Table I.

\begin{itemize}[leftmargin=11pt]
\item 
\textbf{Random Recommendations (RR)}:
Random items are recommended to the users recursively to collect their clicks $\{ \boldsymbol{r}_i \}_{1 \leq i \leq N}$.
The anomalous users are selected according to the anomaly score $\phi (\boldsymbol{r}_i; \{ \boldsymbol{r}_j \}_{1 \leq j \leq N})$ of each user for $1 \leq i \leq N$, where we adopt $\phi (x_i; \{ x_j \}_{1 \leq j \leq N})$ from OCSVM~\cite{Manevitz2002}, LOF~\cite{Breunig2000} and $i$Forest~\cite{Liu2012}.

\item
\textbf{Round-Robin Algorithm (RRA)}:
RRA~\cite{zhuang2017identifying} aims to detect anomalous arms in a multi-armed bandit. 
In our experiment, each user is formulated as one arm of the multi-armed bandit, and RRA is employed to detect anomalous users according to their clicks on random recommended items.

\end{itemize}

% \vspace{-2mm}
\subsubsection{\Algnameabbr{}-R compared with Baselines}

Table~\ref{tb:virtual_taobao_exp_results} shows the ROC-AUC  of \Algnameabbr{}-R and baseline methods with different ratios of fake clicks $a \%$ by the anomalous users. 
We make two observations as follows. 
First, \Algnameabbr{}-R significantly outperforms RRA in all the settings. 
A possible explanation is that RRA ignores each user's state transitions during the interactions and identifies anomalous users according to their average clicking rates on the recommended items.
Thus, it fails to isolate those with fake clicks but without significant change of average clicking rates.
Second, \Algnameabbr{}-R outperforms RR, which verifies that the recommended items generated by the policy are better than random patterns.
Furthermore, as $a \%$ reduces from $1.5\%$ to $0.5\%$, indicating that anomalies behave gradually like normal users, all the baseline methods have significant performance degradation, while \Algnameabbr{}-R shows the least degradation compared with baselines.
% ($0.0015$ of ROC-AUC) ($0.12$ of RR-$i$Forest, $0.05$ of RR-LOF, $0.08$ of RR-OCSVM, and $0.09$ of RRA)

\begin{table}[]
\caption{Hyperparameters of \Algnameabbr{}-R.}
\vspace{-2mm}
\centering
\begin{tabular}{l|c}
  \hline
  Hyperparameter & Value \\
  \hline
  State space $\mathcal{S}$ & $\mathbb{R}^{91}$ \\
  Activation space $\mathcal{A}$ & $\mathbb{R}^{26}$ \\
  Reward space & $\mathbb{R}$ \\
  Max time step $T$ & $10$ \\
  LSTM hidden dim & $256$ \\
  Learning rate & $0.0001$ \\
  Embedding dim $D$ & $10$ \\ 
  Mini-batch size $B$ & $32$ \\
  Replay buffer size & $10^6$ \\
  \hline
\end{tabular}
\label{tb:hyperparameters_SEN/R}
\end{table}

\begin{table} \small
\centering
\caption{ROC-AUC of detecting anomalous users, where anomalous users give fake clicks on $a \%$ of items.}
\vspace{-2mm}
\begin{tabular}{l|c|c|c}
  \hline
    % $a \%$
    \diagbox [width=5em,trim=l] {Detector}{$a \%$}
    & 0.5$\%$ % 1/200
    & 1$\%$ % 2/200
    & 1.5$\%$ \\ % 3/200
  \hline % 1000 step + OCSVM
  RR-$i$Forest & 0.702 $\!\!\pm\!\!$ 0.026 & 0.788 $\!\!\pm\!\!$ 0.035 & 0.822 $\!\!\pm\!\!$ 0.041 \\
  RR-LOF & 0.706 $\!\!\pm\!\!$ 0.011 & 0.726 $\!\!\pm\!\!$ 0.006 & 0.755 $\!\!\pm\!\!$ 0.050 \\
  RR-OCSVM & 0.737 $\!\!\pm\!\!$ 0.040 & 0.781 $\!\!\pm\!\!$ 0.047 & 0.816 $\!\!\pm\!\!$ 0.028 \\
  RRA & 0.541 $\!\!\pm\!\!$ 0.031 & 0.611 $\!\!\pm\!\!$ 0.042 & 0.631 $\!\!\pm\!\!$ 0.026 \\
  \textbf{\Algnameabbr{}-R} & \textbf{0.9975} $\!\!\pm\!\!$ 2e$^{-4}$ & \textbf{0.999} $\!\!\pm\!\!$ 1e$^{-4}$ & \textbf{0.999}  $\!\!\pm\!\!$ 2e$^{-4}$ \\
  \hline
\end{tabular}
\label{tb:virtual_taobao_exp_results}
\end{table}

\begin{figure*}
% \setlength{\abovecaptionskip}{-1mm}
% \setlength{\belowcaptionskip}{-5mm}
% \subfigcapskip=-1mm
\centering
\begin{minipage}{0.32\linewidth}
\centering
\subfigure[]{
\centering
\includegraphics[width=1.0\textwidth]{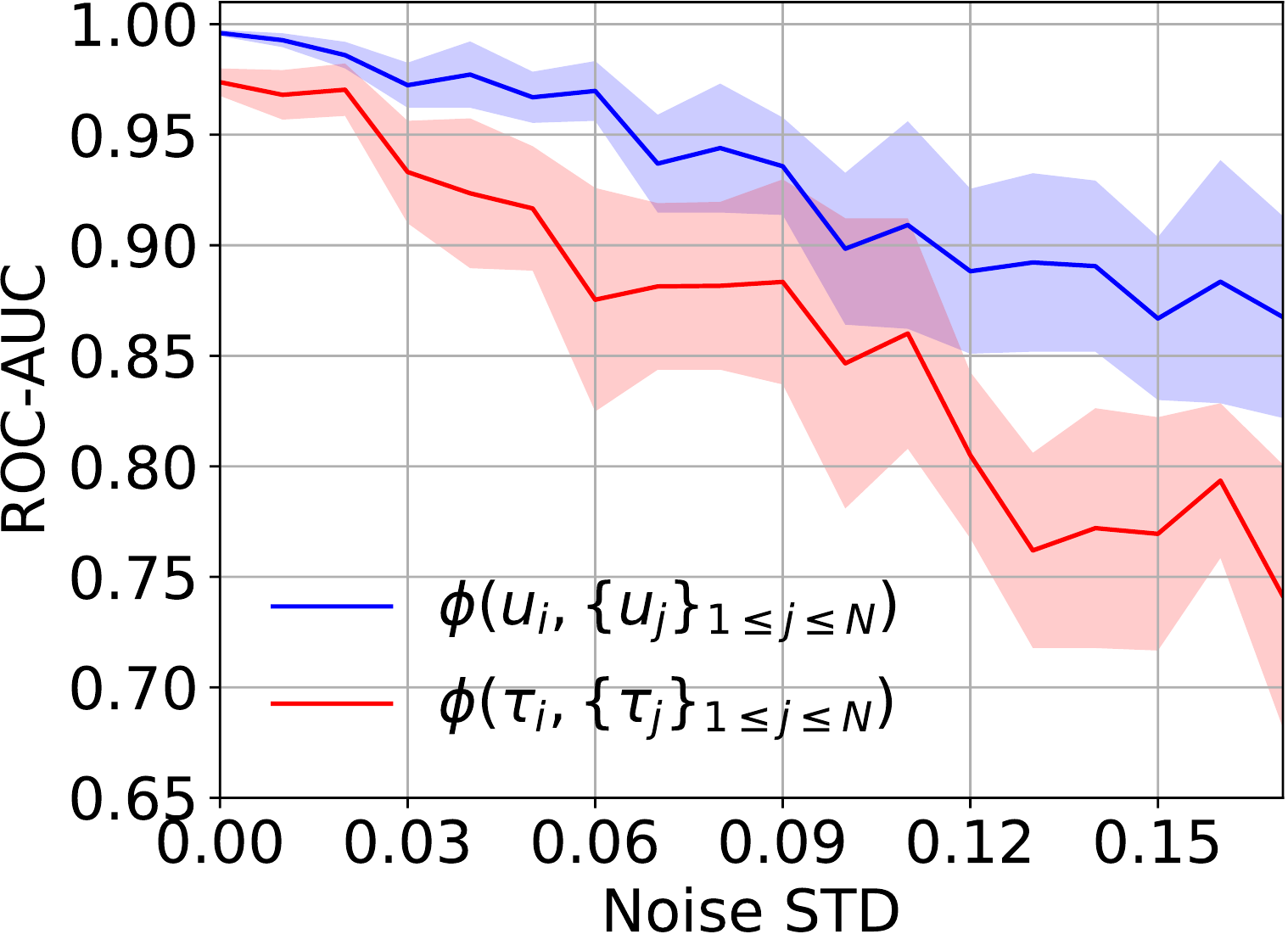}
}
\end{minipage}
\begin{minipage}{0.32\linewidth}
\centering
\subfigure[]{
\centering
\includegraphics[width=1.0\textwidth]{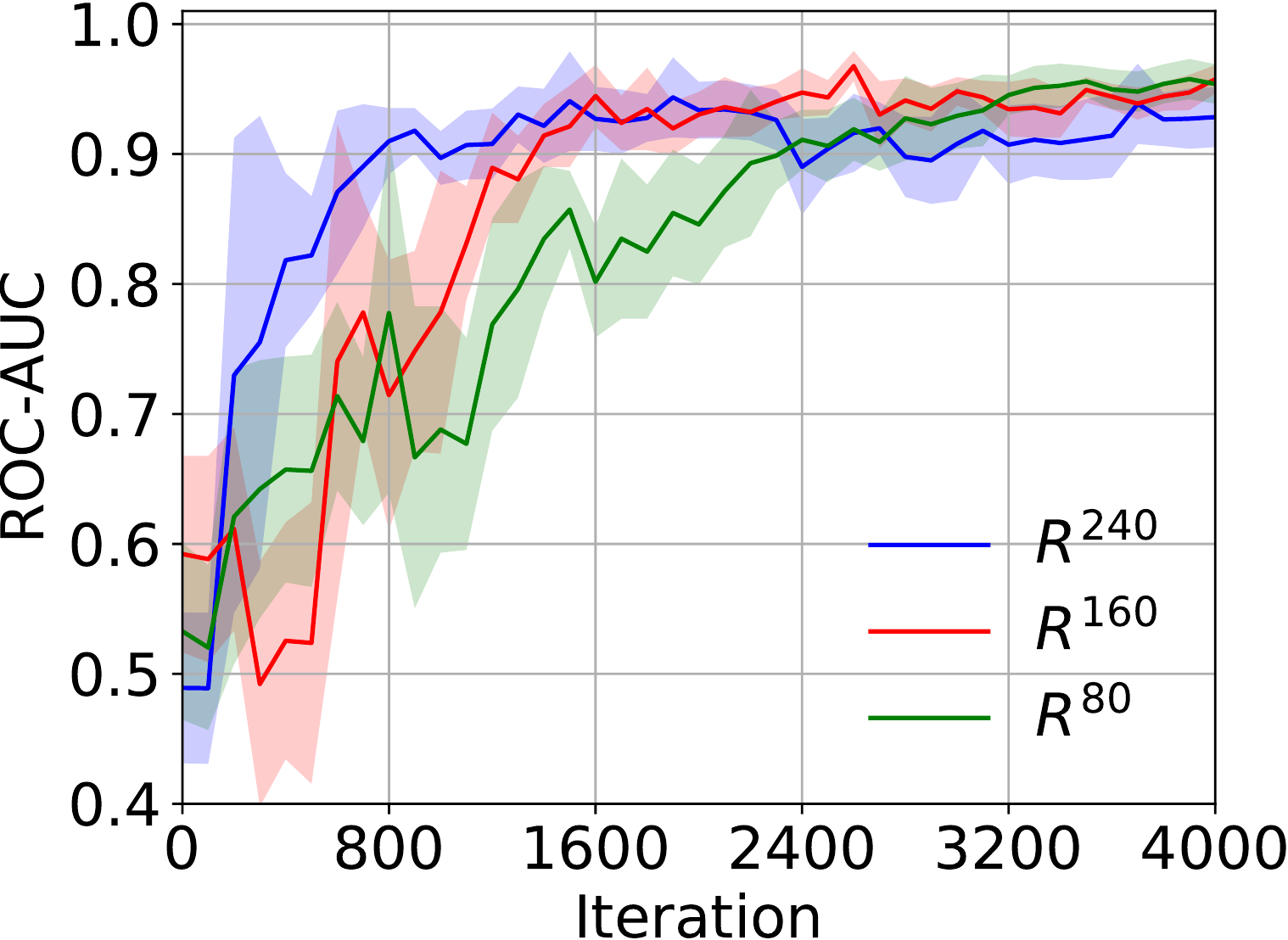}
}
\end{minipage}
% \begin{minipage}{0.32\linewidth}
% \centering
% \subfigure[]{
% \centering
% \includegraphics[width=1.0\textwidth]{figure/halfcheetah_type1_-quiry_time-auc_time-cropped.pdf}
% }
% \end{minipage}
\begin{minipage}{0.32\linewidth}
\centering
\subfigure[]{
\centering
\includegraphics[width=1.0\textwidth]{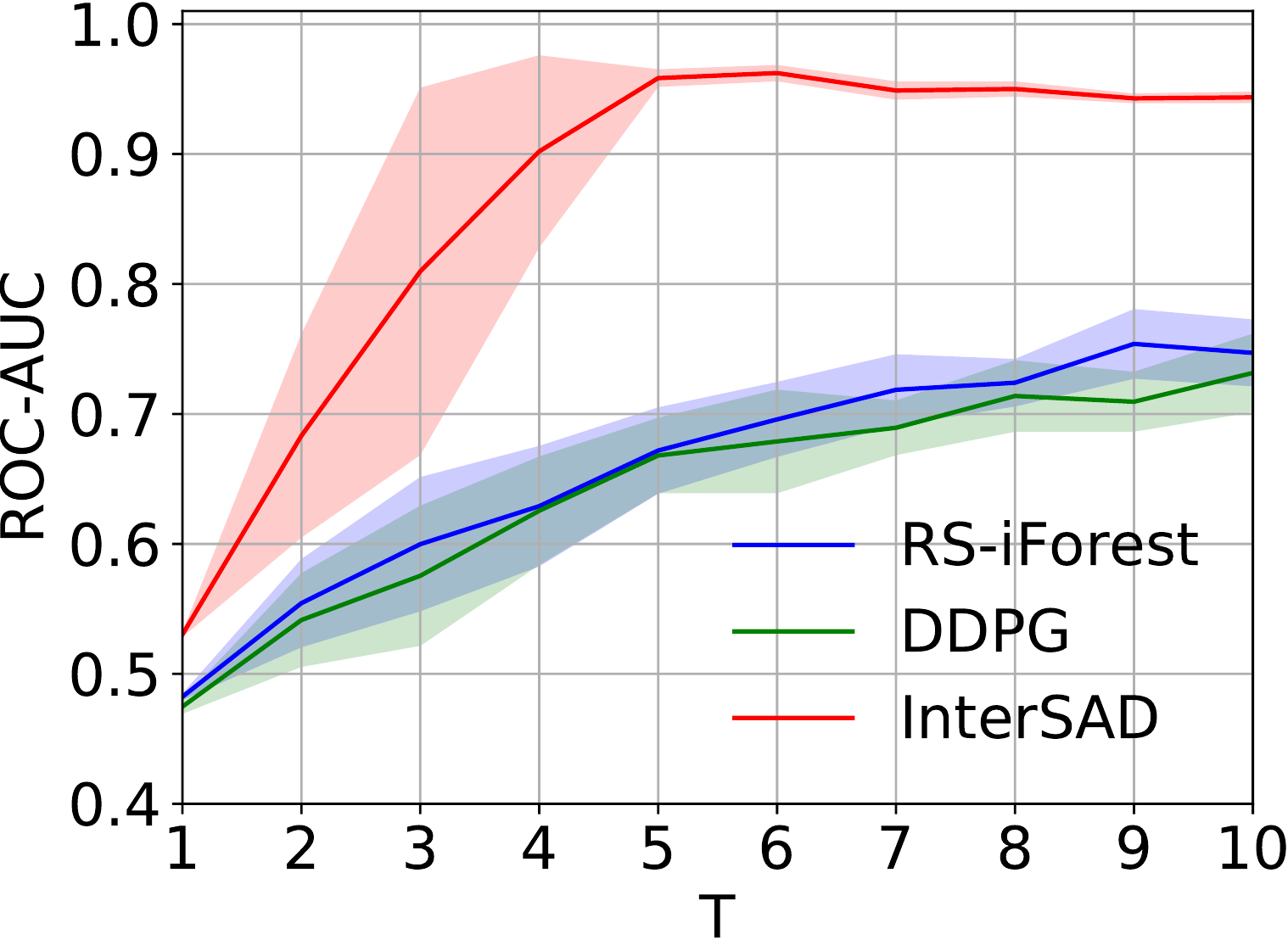}
}
\end{minipage}
\caption{
(a) The detection of anomalous transition system in the trajectory and embedding space versus the intensity of observation noise.
(b) Performance of \Algnameabbr{}-T with different dimensions of embedding space.
(c) Performance of \Algnameabbr{}-T versus the maximum time step $T$ of the interactions.
}
\label{fig:AUC_hyperparmeter}
\end{figure*}

\subsection{Effect of SEN (RQ2)}
\label{exp:2}

To illustrate the effect of SEN, we track the performance of on-going trained policy.
Specifically, we did 50 testings on the $0, 100, \cdots 500$-iteration trained policy, respectively, where different testings are independent with each other.
In each test, we use the trained policy to generate items for randomly chosen $100$ users including $90$ normal and $10$ anomalous users, and collect the trajectories $\{\boldsymbol{\tau}_i\}_{1 \leq i \leq N}$ and rewards $\{\boldsymbol{r}_i\}_{1 \leq i \leq N}$.
We obtain the system embedding $\boldsymbol{u}_i = f_E(\boldsymbol{\tau}_i \mid \theta_E)$ and anomaly score $\phi(\boldsymbol{r}_i; \{\boldsymbol{r}_j\}_{1 \leq j \leq N})$ for each system $\mathcal{M}_i$, and then visualize the relation between $\sum_{i=1}^N ||\boldsymbol{u}_i - \boldsymbol{u}_c||_2$ and the ROC-AUC of detecting anomalous users shown in Figure~\ref{fig:virtual_taobao_2D_embedding}~(a).
A clear inverse correlation between $\sum_{i=1}^N ||\boldsymbol{u}_i - \boldsymbol{u}_c||^2_2$ and ROC-AUC can be observed, where anomaly detection performs better as $\sum_{i=1}^N ||\boldsymbol{u}_i - \boldsymbol{u}_c||^2_2$ reduces.
This relationship certificates the validity of SEN where the policy is updated to minimize $\sum_{i=1}^N ||\boldsymbol{u}_i - \boldsymbol{u}_c||^2_2$ in Equation~(\ref{eq:optimal_mu2}).
Furthermore, the training process of \Algnameabbr{}-R can be tracked by following the arrows in Figure~\ref{fig:virtual_taobao_2D_embedding} (c).
Specifically, as the policy is trained for more iterations, it leads to lower $\sum_{i=1}^N ||\boldsymbol{u}_i - \boldsymbol{u}_c||^2_2$ due to the minimization of Equation~(\ref{eq:optimal_mu2}).
The reduction of $\sum_{i=1}^N ||\boldsymbol{u}_i - \boldsymbol{u}_c||^2_2$ indicates that normal users tends to give consistent clicks, which is beneficial to the anomaly detection as shown in the improvement of ROC-AUC.

% \vspace{-2mm}
% \subsection{Isolation of Anomalous Reward Systems (RQ2)}
% \label{exp:3}

% For the simplicity of expression, the list of items generated by \Algnameabbr{}-R and RR are abbreviated as LI/SEN and LI/RR, respectively.

Another experiment is conducted to illustrate the contribution of SEN to the isolation of anomalous systems.
Specifically, the rewards $\{ \boldsymbol{r}_i \}_{1 \leq i \leq N}$ collected from the interaction with the users are visualized in Figures~\ref{fig:virtual_taobao_2D_embedding} (b) and (c), where $N = 100$ including $90\%$ normal and $10\%$ anomalous users ($a \% = 0.5\%$). The activation for the interaction is generated by \Algnameabbr{}-R and RR for subfigures~(b) and (c), respectively, and $\{ \boldsymbol{r}_i \}_{1 \leq i \leq N}$ is embedded to $\mathbb{R}^2$ by Principal Components Analysis (PCA) for visualization.
According to Figure~\ref{fig:virtual_taobao_2D_embedding} (b), the activation generated by \Algnameabbr{}-R enables the reward collected from normal users to naturally form a cluster, and those of anomalous users can be easily isolated.
In contrast, without SEN in RR, all of the collected rewards from the users spread uniformly in the embedding space, as shown in Figure~\ref{fig:virtual_taobao_2D_embedding} (c), thus leading to worse anomaly detection results than \Algnameabbr{}-R. 

\subsection{Ablation Study (RQ3)}
\label{exp:3}

% remove ablation study of embedding network because the algorithm cannot be trained without embedding network.

To study the necessity of SEN and ERM, we constructed controlled experiments based on 3 candidate models: \Algnameabbr{}-T, \Algnameabbr{}-T without the policy updated by SEN, and \Algnameabbr{}-T without ERM.
They are trained and tested in the experiments of detecting A-\emph{HalfCheetah} and A-\emph{Walker2D} in the same conditions.
We make the observations according to their performance given in Figure~\ref{fig:ablation_results} (b) and (c). First, considerable performance degradation can be observed when removing the policy from \Algnameabbr{}-T, which indicates SEN contributes to distinguishing the anomalous systems. % the learned policy that can minimize Equation~(\ref{eq:optimal_mu2})
Besides, without ERM, \Algnameabbr{}-T not only converges slower but also has noticeable performance degradation after the convergence as well, which implies ERM can stablize the training process by updating the embedding network based on the mixture of trajectories collected in different time stage.

% To study the necessity of policy and embedding network, we remove either of them from \Algnameabbr{}-T for controlled experiments.
% Specifically, \Algnameabbr{}-T without the embedding network (original trajectory as the representation of each system) and that without policy (actions are randomly generated) are tested in the same condition with complete \Algnameabbr{}-T in the experiment of detecting A-\emph{HalfCheetah}.
% As shown in Figure~\ref{fig:ablation_results} (b), we observe considerable performance degradation when either the policy or embedding network is removed, which verifies their necessities in the detection of anomalous systems. 

% Furthermore, without STU, in the training of \Algnameabbr{}-T, the ongoing updated policy is involved in the interaction with each system, which contributes to the failure of convergence.
% Hence, STU can stablize the training process by using the target policy to interacting with each system and to track the ongoing learned policy.

% In addition, \Algnameabbr{}-T without the embedding network still outperforms most of the baselines, according to Table~\ref{tb:gym_exp_result}. 
% Considering LSTM Autoencoder's expense, in mobile devices with finite memory, \Algnameabbr{}-T without embedding network can be a compromised solution to significantly reduce storage cost with satisfactory performance.

% Anomaly detection depends on the original trajectories for \Algnameabbr{}-T without the embedding network; and relies on the embedded trajectories generated by random sampling for \Algnameabbr{}-T without policy. 

\subsection{Anomaly Detection in the embedding space (RQ4)}
\label{exp:4}

We study using system embedding as an alternative option of estimating anomaly scores in detecting of anomalous transition systems.
Specifically, we consider that interactions with the systems are distorted by the observation noise $\boldsymbol{\tau}_i = [ s_{i,0} + n_{i,0}, a_{i,0}, \cdots, s_{i,T} + n_{i,T} ]$, where $a_{i,t} = \mu(s_{i,t} + n_{i,t} \mid \theta_{\mu})$ and $n_{i,t} \sim \mathcal{N}(0, \sigma^2)$ for $0 \leq t \leq T$ and $1\leq i\leq N$.
The performance of isolating the anomaly transition systems in the trajectory and embedding space is given in Figure~\ref{fig:AUC_hyperparmeter}~(a), where the anomaly score of system $\mathcal{M}_i$ is estimated by $\phi(\boldsymbol{u}_i; \{ \boldsymbol{u}_j \}_{1\leq j\leq N})$ and $\phi(\boldsymbol{\tau}_i; \{ \boldsymbol{\tau}_j \}_{1\leq j\leq N})$ for the blue and red curves, respectively.
According to Figure~\ref{fig:AUC_hyperparmeter}~(a), better performance is achieved by conducting anomaly detection in the embedding space, especially for the case where the observation noise is more significant.
The possible reason is that estimating anomaly scores in the embedding space leads to more robust detection against the observation noise, due to the fact that the encoder $f_E (\boldsymbol{\tau} \mid \theta_E)$ learns more abstract representations for the systems in the training of \Algnameabbr{}-T, which can reduce the influence of noise in the testing stage.

% \vspace{-2mm}
\subsection{Analysis of Hyperparameters (RQ5)}
\label{exp:5}

We study the impact of the hyperparameters, including embedding dimension $D$ and the maximum time step of interaction $T$ in Figures~\ref{fig:AUC_hyperparmeter}~(b) and (c), respectively.
According to Figure~\ref{fig:AUC_hyperparmeter}~(b), the ROC-AUC converges slower as the embedding space reduces from $\mathbb{R}^{240} \text{ to } \mathbb{R}^{80}$, since it needs more iterations to learn lower dimensional embedding from original trajectory space $\mathbb{R}^{320}$. 
However, lower-dimensional embedded trajectories provide better abstractions for systems.
Hence, a slightly higher ROC-AUC of $\mathbb{R}^{80}$ (green curve) than $\mathbb{R}^{160}$ and $\mathbb{R}^{240}$ can be observed in the final iterations.
In addition, according to Figure~\ref{fig:AUC_hyperparmeter}~(c), all methods achieve better ROC-AUC as $T$ grows due to the fact that longer trajectories help produce more integrated representation for each system.
Among them, \Algnameabbr{}-T keeps ROC-AUC more than $0.95$ when $T > 4$, which indicates its more consistent performance than the baselines, and its superiority in real-world tasks considering the cost of collecting trajectories from real systems.

% According to Figures~\ref{fig:AUC_hyperparmeter} (b) and (c), all methods achieve better ROC-AUC with either larger collected number or longer trajectories because more trajectories can reflect the statistical characteristic of each system more accurately, and longer trajectories can provide more integrated representation for each system.
% Among them, \Algnameabbr{}-T keeps ROC-AUC more than $0.95$ when sampling number $>3$ or maximum time step $>5$, indicating its consistent performance versus these two hyperparameters. 
% Besides, compared with baselines in Figures~\ref{fig:AUC_hyperparmeter} (a) and (b), where anomaly detection depends on few and short trajectories, \Algnameabbr{}-T's outperformance indicates its superiority in real-world tasks considering the budget of collecting trajectories from each system.

\section{Related Work}

Anomaly detection refers to the problem of detecting anomalous instances that hardly conform to the patterns of normal instances \cite{chandola2009anomaly}. 
% Existing works on anomaly detection can be categorized into unsupervised, supervised and semi-supervised learning algorithms.
Unsupervised anomaly detection has been proven to be an effective way in the scenario where normal instances are far more frequent than anomalies~\cite{eskin2002geometric}.
For example, One class SVM (OCSVM) identifies an optimal hyper-sphere characterized by a radius and a center to cover normal instances and exclude anomalies~\cite{Manevitz2002, ruff2018deep};
Local Oulier Factor (LOF) estimates the density of the nearest neighbors for each instance, and regards the instances with low density as anomalies~\cite{Breunig2000};
Isolation Forest ($i$Forest) recursively generates segmentation on the instances by randomly selecting an attribute and generating splits for the chosen attribute~\cite{Liu2012}.
% deep learning method
Recently, many deep anomaly detection algorithms have been proposed to deal with complex data structures, such as images, graphs and time-series~\cite{chalapathy2019deep,ruff2019deep}.
For example, Autoencoder, which estimates the anomaly scores based on reconstruction errors~\cite{zhou2017anomaly, hendrycks2016baseline, zhao2017spatio}, is widely adopted for image anomaly detection;
Generative Adversarial Networks leverage data argumentation to address the data imbalance issue between normal and anomalous samples~\cite{deecke2018image, schlegl2017unsupervised, li2018anomaly};
LSTM plays an important role in time-series anomaly detection data~\cite{bontemps2016collective, malhotra2016lstm, malhotra2015long, ergen2019unsupervised, lai2020tods}.
Other research focuses on active anomaly detection~\cite{zha2020meta}, multiple semantic embedding~\cite{shalev2018out}, uncertainty estimation and interpretation~\cite{depeweg2018decomposition, sipple2020interpretable, blundell2015weight,liu2017contextual}, etc.

% interactive anomaly detection
While the existing studies have proposed various anomaly detection algorithms to tackle different types of data, they often assume the availability of a static dataset and are inapplicable for system-wise anomaly detection. This is because a system could not be explicitly observed as data unless an activation signal is provided and often requires  appropriate interactions to identify abnormal responses. To the best of our knowledge, the only research line that considers interactions is anomaly detection on multi-armed bandits~\cite{zhuang2017identifying, ban2020generic}. However, multi-armed bandits cannot model the complex state transitions in a system. Whereas, we formulate a system as a Markov decision process, which naturally models the state transitions during interactions~\cite{todorov2012mujoco, shani2005mdp, shi2019virtual}. Based on this formulation, we propose an end-to-end framework for system-wise anomaly detection and demonstrate its effectiveness on systems with very complex state transitions.
\section{Conclusions and Future Work}
In this work, we propose to investigate system-wise anomaly detection. Unlike many traditional anomaly detection scenarios that assume the datasets are given, the characteristics of systems are not readily observed as data. This poses new challenges, such as how to model a system and formulate the anomalies, and how to find the  effective activation signals. To address these challenges, we first use Markov decision process to describe interactive systems and give a formal definition of anomalies. Then, we propose \Algnameabbr{} for end-to-end system-wise anomaly detection, which includes i) a system embedding neutralization module to encourage consistent behaviors of normal systems for the isolation of anomalous systems, and ii) an experience replay mechanism to stabilize training.
Experimental results demonstrate the effectiveness of \Algnameabbr{} in identifying anomalous robotic systems and detecting malicious attacks in recommender systems.
In the future, we will study more sophisticated systems.

%In this work, we conduct a pilot study on system-wise anomaly detection, which involves: 
%(i) using Markov decision process to characterize interactive systems, and giving the formal definition of anomalous systems;
%(ii) proposing \Algnameabbr{} for end-to-end system-wise anomaly detection which involves system embedding neutralization to encourage consistent behaviors of normal systems for the isolation of anomalous systems;
%(iii) adopting experience replay mechanism to stabilize the training procedure during the real-time interactions with the systems.
%Experimental results demonstrate the effectiveness of \Algnameabbr{} in the scenario of identifying anomalous robotic systems and detecting malicious attacks in recommender systems.
%In the future, we will explore the possibility of extending \Algnameabbr{} to more sophisticated scenarios.

% \section*{Acknowledgment}

% The preferred spelling of the word ``acknowledgment'' in America is without 
% an ``e'' after the ``g''. Avoid the stilted expression ``one of us (R. B. 
% G.) thanks $\ldots$''. Instead, try ``R. B. G. thanks$\ldots$''. Put sponsor 
% acknowledgments in the unnumbered footnote on the first page.

% \bibliographystyle{named}
\bibliographystyle{IEEEtran}
\bibliography{adrl}

% Generated by IEEEtran.bst, version: 1.14 (2015/08/26)
\begin{thebibliography}{10}
\providecommand{\url}[1]{#1}
\csname url@samestyle\endcsname
\providecommand{\newblock}{\relax}
\providecommand{\bibinfo}[2]{#2}
\providecommand{\BIBentrySTDinterwordspacing}{\spaceskip=0pt\relax}
\providecommand{\BIBentryALTinterwordstretchfactor}{4}
\providecommand{\BIBentryALTinterwordspacing}{\spaceskip=\fontdimen2\font plus
\BIBentryALTinterwordstretchfactor\fontdimen3\font minus
  \fontdimen4\font\relax}
\providecommand{\BIBforeignlanguage}[2]{{%
\expandafter\ifx\csname l@#1\endcsname\relax
\typeout{** WARNING: IEEEtran.bst: No hyphenation pattern has been}%
\typeout{** loaded for the language `#1'. Using the pattern for}%
\typeout{** the default language instead.}%
\else
\language=\csname l@#1\endcsname
\fi
#2}}
\providecommand{\BIBdecl}{\relax}
\BIBdecl

\bibitem{ahmed2016survey}
M.~Ahmed, A.~N. Mahmood, and M.~R. Islam, ``A survey of anomaly detection
  techniques in financial domain,'' \emph{Future Generation Computer Systems},
  vol.~55, pp. 278--288, 2016.

\bibitem{garcia2009anomaly}
P.~Garcia-Teodoro, J.~Diaz-Verdejo, G.~Maci{\'a}-Fern{\'a}ndez, and
  E.~V{\'a}zquez, ``Anomaly-based network intrusion detection: Techniques,
  systems and challenges,'' \emph{computers \& security}, vol.~28, no. 1-2, pp.
  18--28, 2009.

\bibitem{fu2006finding}
A.~W.-C. Fu, O.~T.-W. Leung, E.~Keogh, and J.~Lin, ``Finding time series
  discords based on haar transform,'' in \emph{International Conference on
  Advanced Data Mining and Applications}.\hskip 1em plus 0.5em minus
  0.4em\relax Springer, 2006, pp. 31--41.

\bibitem{keogh2002finding}
E.~Keogh, S.~Lonardi, and B.-c. Chiu, ``Finding surprising patterns in a time
  series database in linear time and space,'' in \emph{Proceedings of the
  eighth ACM SIGKDD international conference on Knowledge discovery and data
  mining}, 2002, pp. 550--556.

\bibitem{yankov2008disk}
D.~Yankov, E.~Keogh, and U.~Rebbapragada, ``Disk aware discord discovery:
  Finding unusual time series in terabyte sized datasets,'' \emph{Knowledge and
  Information Systems}, vol.~17, no.~2, pp. 241--262, 2008.

\bibitem{zhao2019pyod}
Y.~Zhao, Z.~Nasrullah, and Z.~Li, ``Pyod: A python toolbox for scalable outlier
  detection,'' \emph{arXiv preprint arXiv:1901.01588}, 2019.

\bibitem{lai2020tods}
K.-H. Lai, D.~Zha, G.~Wang, J.~Xu, Y.~Zhao, D.~Kumar, Y.~Chen, P.~Zumkhawaka,
  M.~Wan, D.~Martinez \emph{et~al.}, ``Tods: An automated time series outlier
  detection system,'' \emph{arXiv preprint arXiv:2009.09822}, 2020.

\bibitem{li2020pyodds}
Y.~Li, D.~Zha, P.~Venugopal, N.~Zou, and X.~Hu, ``Pyodds: An end-to-end outlier
  detection system with automated machine learning,'' in \emph{Companion
  Proceedings of the Web Conference 2020}, 2020, pp. 153--157.

\bibitem{chalapathy2019deep}
R.~Chalapathy and S.~Chawla, ``Deep learning for anomaly detection: A survey,''
  \emph{arXiv preprint arXiv:1901.03407}, 2019.

\bibitem{brockman2016openai}
G.~Brockman, V.~Cheung, L.~Pettersson, J.~Schneider, J.~Schulman, J.~Tang, and
  W.~Zaremba, ``Openai gym,'' \emph{arXiv preprint arXiv:1606.01540}, 2016.

\bibitem{shi2019virtual}
J.-C. Shi, Y.~Yu, Q.~Da, S.-Y. Chen, and A.-X. Zeng, ``Virtual-taobao:
  Virtualizing real-world online retail environment for reinforcement
  learning,'' in \emph{Proceedings of the AAAI Conference on Artificial
  Intelligence}, vol.~33, 2019, pp. 4902--4909.

\bibitem{wilson2013power}
D.~C. Wilson and C.~E. Seminario, ``When power users attack: assessing impacts
  in collaborative recommender systems,'' in \emph{Proceedings of the 7th ACM
  conference on Recommender systems}, 2013, pp. 427--430.

\bibitem{aktukmak2019quick}
M.~Aktukmak, Y.~Yilmaz, and I.~Uysal, ``Quick and accurate attack detection in
  recommender systems through user attributes,'' in \emph{Proceedings of the
  13th ACM Conference on Recommender Systems}, 2019, pp. 348--352.

\bibitem{zheng2018drn}
G.~Zheng, F.~Zhang, Z.~Zheng, Y.~Xiang, N.~J. Yuan, X.~Xie, and Z.~Li, ``Drn: A
  deep reinforcement learning framework for news recommendation,'' in
  \emph{Proceedings of the 2018 World Wide Web Conference}, 2018, pp. 167--176.

\bibitem{kober2013reinforcement}
J.~Kober, J.~A. Bagnell, and J.~Peters, ``Reinforcement learning in robotics: A
  survey,'' \emph{The International Journal of Robotics Research}, vol.~32,
  no.~11, pp. 1238--1274, 2013.

\bibitem{kriegel2008angle}
H.-P. Kriegel, M.~Schubert, and A.~Zimek, ``Angle-based outlier detection in
  high-dimensional data,'' in \emph{Proceedings of the 14th ACM SIGKDD
  international conference on Knowledge discovery and data mining}, 2008, pp.
  444--452.

\bibitem{Liu2012}
F.~T. Liu, K.~Ting, and Z.-H. Zhou, ``Isolation-based anomaly detection,''
  \emph{ACM Transactions on Knowledge Discovery From Data - TKDD}, vol.~6, pp.
  1--39, 03 2012.

\bibitem{Breunig2000}
M.~Breunig, H.-P. Kriegel, R.~Ng, and J.~Sander, ``Lof: Identifying
  density-based local outliers.'' in \emph{ACM Sigmod Record}, vol.~29, 06
  2000, pp. 93--104.

\bibitem{Manevitz2002}
L.~Manevitz and M.~Yousef, ``One-class svms for document classification,''
  \emph{J. Mach. Learn. Res}, vol.~2, pp. 139--154, 01 2002.

\bibitem{lin1992self}
L.-J. Lin, ``Self-improving reactive agents based on reinforcement learning,
  planning and teaching,'' \emph{Machine learning}, vol.~8, no. 3-4, pp.
  293--321, 1992.

\bibitem{zha2019experience}
D.~Zha, K.-H. Lai, K.~Zhou, and X.~Hu, ``Experience replay optimization,'' in
  \emph{International Joint Conference on Artificial Intelligence}, 2019.

\bibitem{lillicrap2015continuous}
T.~P. Lillicrap, J.~J. Hunt, A.~Pritzel, N.~Heess, T.~Erez, Y.~Tassa,
  D.~Silver, and D.~Wierstra, ``Continuous control with deep reinforcement
  learning,'' \emph{arXiv preprint arXiv:1509.02971}, 2015.

\bibitem{yang2017detecting}
Z.~Yang and Z.~Cai, ``Detecting abnormal profiles in collaborative filtering
  recommender systems,'' \emph{Journal of Intelligent Information Systems},
  vol.~48, no.~3, pp. 499--518, 2017.

\bibitem{seminario2014attacking}
C.~E. Seminario and D.~C. Wilson, ``Attacking item-based recommender systems
  with power items,'' in \emph{Proceedings of the 8th ACM Conference on
  Recommender systems}, 2014, pp. 57--64.

\bibitem{zhuang2017identifying}
H.~Zhuang, C.~Wang, and Y.~Wang, ``Identifying outlier arms in multi-armed
  bandit,'' in \emph{Advances in Neural Information Processing Systems}, 2017,
  pp. 5204--5213.

\bibitem{chandola2009anomaly}
V.~Chandola, A.~Banerjee, and V.~Kumar, ``Anomaly detection: A survey,''
  \emph{ACM computing surveys (CSUR)}, vol.~41, no.~3, pp. 1--58, 2009.

\bibitem{eskin2002geometric}
E.~Eskin, A.~Arnold, M.~Prerau, L.~Portnoy, and S.~Stolfo, ``A geometric
  framework for unsupervised anomaly detection,'' in \emph{Applications of data
  mining in computer security}.\hskip 1em plus 0.5em minus 0.4em\relax
  Springer, 2002, pp. 77--101.

\bibitem{ruff2018deep}
L.~Ruff, R.~Vandermeulen, N.~Goernitz, L.~Deecke, S.~A. Siddiqui, A.~Binder,
  E.~M{\"u}ller, and M.~Kloft, ``Deep one-class classification,'' in
  \emph{ICML}, 2018.

\bibitem{ruff2019deep}
L.~Ruff, R.~A. Vandermeulen, N.~G{\"o}rnitz, A.~Binder, E.~M{\"u}ller, K.-R.
  M{\"u}ller, and M.~Kloft, ``Deep semi-supervised anomaly detection,''
  \emph{arXiv preprint arXiv:1906.02694}, 2019.

\bibitem{zhou2017anomaly}
C.~Zhou and R.~C. Paffenroth, ``Anomaly detection with robust deep
  autoencoders,'' in \emph{Proceedings of the 23rd ACM SIGKDD international
  conference on knowledge discovery and data mining}, 2017, pp. 665--674.

\bibitem{hendrycks2016baseline}
D.~Hendrycks and K.~Gimpel, ``A baseline for detecting misclassified and
  out-of-distribution examples in neural networks,'' \emph{arXiv preprint
  arXiv:1610.02136}, 2016.

\bibitem{zhao2017spatio}
Y.~Zhao, B.~Deng, C.~Shen, Y.~Liu, H.~Lu, and X.-S. Hua, ``Spatio-temporal
  autoencoder for video anomaly detection,'' in \emph{Proceedings of the 25th
  ACM international conference on Multimedia}, 2017, pp. 1933--1941.

\bibitem{deecke2018image}
L.~Deecke, R.~Vandermeulen, L.~Ruff, S.~Mandt, and M.~Kloft, ``Image anomaly
  detection with generative adversarial networks,'' in \emph{Joint european
  conference on machine learning and knowledge discovery in databases}.\hskip
  1em plus 0.5em minus 0.4em\relax Springer, 2018, pp. 3--17.

\bibitem{schlegl2017unsupervised}
T.~Schlegl, P.~Seeb{\"o}ck, S.~M. Waldstein, U.~Schmidt-Erfurth, and G.~Langs,
  ``Unsupervised anomaly detection with generative adversarial networks to
  guide marker discovery,'' in \emph{International conference on information
  processing in medical imaging}.\hskip 1em plus 0.5em minus 0.4em\relax
  Springer, 2017, pp. 146--157.

\bibitem{li2018anomaly}
D.~Li, D.~Chen, J.~Goh, and S.-k. Ng, ``Anomaly detection with generative
  adversarial networks for multivariate time series,'' \emph{arXiv preprint
  arXiv:1809.04758}, 2018.

\bibitem{bontemps2016collective}
L.~Bontemps, J.~McDermott, N.-A. Le-Khac \emph{et~al.}, ``Collective anomaly
  detection based on long short-term memory recurrent neural networks,'' in
  \emph{International Conference on Future Data and Security
  Engineering}.\hskip 1em plus 0.5em minus 0.4em\relax Springer, 2016, pp.
  141--152.

\bibitem{malhotra2016lstm}
P.~Malhotra, A.~Ramakrishnan, G.~Anand, L.~Vig, P.~Agarwal, and G.~Shroff,
  ``Lstm-based encoder-decoder for multi-sensor anomaly detection,''
  \emph{arXiv preprint arXiv:1607.00148}, 2016.

\bibitem{malhotra2015long}
P.~Malhotra, L.~Vig, G.~Shroff, and P.~Agarwal, ``Long short term memory
  networks for anomaly detection in time series,'' in \emph{Proceedings},
  vol.~89.\hskip 1em plus 0.5em minus 0.4em\relax Presses universitaires de
  Louvain, 2015, pp. 89--94.

\bibitem{ergen2019unsupervised}
T.~Ergen and S.~S. Kozat, ``Unsupervised anomaly detection with lstm neural
  networks,'' \emph{IEEE transactions on neural networks and learning systems},
  vol.~31, no.~8, pp. 3127--3141, 2019.

\bibitem{zha2020meta}
D.~Zha, K.-H. Lai, M.~Wan, and X.~Hu, ``Meta-aad: Active anomaly detection with
  deep reinforcement learning,'' \emph{arXiv preprint arXiv:2009.07415}, 2020.

\bibitem{shalev2018out}
G.~Shalev, Y.~Adi, and J.~Keshet, ``Out-of-distribution detection using
  multiple semantic label representations,'' \emph{arXiv preprint
  arXiv:1808.06664}, 2018.

\bibitem{depeweg2018decomposition}
S.~Depeweg, J.-M. Hernandez-Lobato, F.~Doshi-Velez, and S.~Udluft,
  ``Decomposition of uncertainty in bayesian deep learning for efficient and
  risk-sensitive learning,'' in \emph{International Conference on Machine
  Learning}.\hskip 1em plus 0.5em minus 0.4em\relax PMLR, 2018, pp. 1184--1193.

\bibitem{sipple2020interpretable}
J.~Sipple, ``Interpretable, multidimensional, multimodal anomaly detection with
  negative sampling for detection of device failure,'' in \emph{International
  Conference on Machine Learning}.\hskip 1em plus 0.5em minus 0.4em\relax PMLR,
  2020, pp. 9016--9025.

\bibitem{blundell2015weight}
C.~Blundell, J.~Cornebise, K.~Kavukcuoglu, and D.~Wierstra, ``Weight
  uncertainty in neural network,'' in \emph{International Conference on Machine
  Learning}.\hskip 1em plus 0.5em minus 0.4em\relax PMLR, 2015, pp. 1613--1622.

\bibitem{liu2017contextual}
N.~Liu, D.~Shin, and X.~Hu, ``Contextual outlier interpretation,'' \emph{arXiv
  preprint arXiv:1711.10589}, 2017.

\bibitem{ban2020generic}
Y.~Ban and J.~He, ``Generic outlier detection in multi-armed bandit,'' in
  \emph{Proceedings of the 26th ACM SIGKDD International Conference on
  Knowledge Discovery \& Data Mining}, 2020, pp. 913--923.

\bibitem{todorov2012mujoco}
E.~Todorov, T.~Erez, and Y.~Tassa, ``Mujoco: A physics engine for model-based
  control,'' in \emph{2012 IEEE/RSJ International Conference on Intelligent
  Robots and Systems}.\hskip 1em plus 0.5em minus 0.4em\relax IEEE, 2012, pp.
  5026--5033.

\bibitem{shani2005mdp}
G.~Shani, D.~Heckerman, R.~I. Brafman, and C.~Boutilier, ``An mdp-based
  recommender system.'' \emph{Journal of Machine Learning Research}, vol.~6,
  no.~9, 2005.

\end{thebibliography}

\section{Appendix}

\subsection{Proof of $\sum_{i=1}^N \sum_{j=1}^N ||\textbf{u}_i \!-\! \textbf{u}_j||_2 \!\leq\! 2(N\!-\!1)\! \sum_{i=1}^N ||\textbf{u}_i \!-\! \textbf{u}_c||_2$ in Section~\ref{sec:32}}

For any vectors $\boldsymbol{x}$ and $\boldsymbol{y} \in \mathbb{R}^N$, we have:
\begin{equation}
\begin{aligned}
||\boldsymbol{x} - \boldsymbol{y}||_2 
&= \sqrt{\sum_{i=1} x_i^2 + y_i^2 - 2 x_i y_i}
\\
&= \sqrt{\sum_{i=1} x_i^2 + \sum_{i=1} y_i^2 +  2 \sum_{i=1} x_i (-y_i)}.
\nonumber
\end{aligned}
\end{equation}
According to the Cauchy-Schwarz inequality $\forall a_i, b_i \!\in\! \mathbb{R}$, $(\sum_{i=1} a_i b_i)^2 \leq \sum_{i=1} a_i^2 \sum_{i=1} b_i^2$, we have:
\begin{equation}
\begin{aligned}
||\boldsymbol{x} - \boldsymbol{y}||_2 
&\leq \sqrt{\sum_{i=1} x_i^2 + \sum_{i=1} y_i^2 +  2 \sqrt{\sum_{i=1} x_i^2 \sum_{i=1} y_i^2}}
\\
&= \sqrt{\sum_{i=1} x_i^2} + \sqrt{\sum_{i=1} y_i^2}
\\
&= ||\boldsymbol{x}||_2 + ||\boldsymbol{y}||_2.
\nonumber
\end{aligned}
\end{equation}
Let $\boldsymbol{x} = \boldsymbol{u}_i - \boldsymbol{u}_c$ and $\boldsymbol{y} = \boldsymbol{u}_j - \boldsymbol{u}_c$, thus we have:
\begin{equation}
\label{eq:triangle_inequality}
||\boldsymbol{u}_i - \boldsymbol{u}_j||_2 \leq ||\boldsymbol{u}_i - \boldsymbol{u}_c||_2 + ||\boldsymbol{u}_j - \boldsymbol{u}_c||_2.
\end{equation}
Plugging the inequality into $\sum_{i=1}^N \sum_{j=1}^N || \boldsymbol{u}_i - \boldsymbol{u}_j ||_2$, we have:
\begin{equation}
\begin{aligned}
&\sum_{i=1}^N \sum_{j=1}^N || \boldsymbol{u}_i \!-\! \boldsymbol{u}_j ||_2 = \sum_{i=1}^N \sum_{j\neq k} || \boldsymbol{u}_i \!-\! \boldsymbol{u}_j ||_2
\\
&\leq \sum_{i=1}^N \sum_{j=1}^N ||\boldsymbol{u}_i - \boldsymbol{u}_c||_2 + ||\boldsymbol{u}_j - \boldsymbol{u}_c||_2
\\
&= \sum_{i=1}^N \sum_{j\neq k} ||\boldsymbol{u}_i - \boldsymbol{u}_c||_2 + \sum_{j=1}^N \sum_{k \neq j} ||\boldsymbol{u}_j - \boldsymbol{u}_c||_2
\\
&= (N-1)\sum_{i=1}^N ||\boldsymbol{u}_i - \boldsymbol{u}_c||_2 + (N-1) \sum_{j=1}^N ||\boldsymbol{u}_j - \boldsymbol{u}_c||_2
\\
&= 2(N-1)\sum_{i=1}^N ||\boldsymbol{u}_i - \boldsymbol{u}_c||_2.
\nonumber
\end{aligned}
\end{equation}

\end{document}